\documentclass{article}

\usepackage{arxiv}

\usepackage[utf8]{inputenc} % allow utf-8 input
\usepackage[T1]{fontenc}    % use 8-bit T1 fonts
\usepackage{hyperref}       % hyperlinks
\usepackage{url}            % simple URL typesetting
\usepackage{booktabs}       % professional-quality tables
\usepackage{amsfonts}       % blackboard math symbols
\usepackage{nicefrac}       % compact symbols for 1/2, etc.
\usepackage{microtype}      % microtypography
\usepackage{lipsum}		% Can be removed after putting your text content
\usepackage{multirow}
\usepackage{paralist}
\usepackage{graphicx}

\usepackage{doi}
\usepackage{xcolor}

\definecolor{innerboxcolor}{rgb}{.9,.95,1}
\definecolor{outerlinecolor}{rgb}{.6,0,.2}
\definecolor{outerlinecoloreb}{rgb}{0,1,0}
\definecolor{innerboxcoloreb}{rgb}{1,1,1}
\definecolor{OI-purple}{RGB}{204,121,167}
\definecolor{OI-blue}{RGB}{30, 136, 229}

\hypersetup{
    colorlinks=true,
    linkcolor=OI-blue,
    citecolor=OI-blue,
    filecolor=magenta,      
    urlcolor=OI-blue,
    pdftitle={Overleaf Example},
    pdfpagemode=FullScreen,
    }

\usepackage{natbib}
\newcommand{\Description}[1]{\caption{#1}}

\title{A pipeline for enabling path-specific causal fairness in observational health data}

\author{
Aparajita Kashyap \\
Department of Biomedical Informatics\\
Columbia University\\
New York, NY, USA \\
\texttt{ak4885@cumc.columbia.edu}
\And
Sara Matijevic \\
Big Data Institute\\
University of Oxford\\
Oxford, UK \\
\texttt{sm6088@cumc.columbia.edu}
\And
No\'emie Elhadad \\
Department of Biomedical Informatics\\
Columbia University\\
New York, NY, USA \\
\texttt{noemie.elhadad@columbia.edu}
\And
Steven A.~Kushner \\
Department of Psychiatry\\
Columbia University\\
New York, NY, USA \\
\texttt{sk2602@cumc.columbia.edu}
\And
Shalmali Joshi \\
Department of Biomedical Informatics\\
Columbia University\\
New York, NY, USA \\
\texttt{sj3261@cumc.columbia.edu}
}

\renewcommand{\shorttitle}{A pipeline for enabling path-specific causal fairness in observational health data}

\hypersetup{
pdftitle={A template for the arxiv style},
pdfsubject={q-bio.NC, q-bio.QM},
pdfauthor={David S.~Hippocampus, Elias D.~Striatum},
pdfkeywords={Causal fairness, healthcare, foundation models, causal inference, observational health data, fair machine learning},
}

\begin{document}
\maketitle

\begin{abstract}
  When training machine learning (ML) models for potential deployment in a healthcare setting, it is essential to ensure that they do not replicate or exacerbate existing healthcare biases. Although many definitions of fairness exist, we focus on \textit{path-specific causal fairness}, which allows us to better consider the social and medical contexts in which biases occur (e.g., direct discrimination by a clinician or model versus bias due to differential access to the healthcare system) and to characterize how these biases may appear in learned models. In this work, we map the structural fairness model to the observational healthcare setting and create a generalizable pipeline for training causally fair models\footnote{Code available at \url{https://github.com/reAIM-Lab/causal_fairness_health_pipeline}}. The pipeline explicitly considers specific healthcare context and disparities to define a target "fair" model. Our work fills two major gaps: first, we expand on characterizations of the "fairness-accuracy" tradeoff by detangling direct and indirect sources of bias and jointly presenting these fairness considerations alongside considerations of accuracy in the context of broadly known biases. Second, we demonstrate how a foundation model trained without fairness constraints on observational health data can be leveraged to generate causally fair downstream predictions in tasks with known social and medical disparities. This work presents a model-agnostic pipeline for training causally fair machine learning models that address both direct and indirect forms of healthcare bias. 
\end{abstract}

% keywords can be removed

\section{Introduction}
The need for fairness in machine learning (ML) is well-documented, particularly in the context of healthcare, where there are known systemic disparities that can lead to bias in data collection and in downstream ML models \citep{chen_biomedicalfairness_5things_2021}. Bias in the healthcare system exists through several avenues: implicit biases held by clinicians can directly the lower quality of patient care \citep{FitzGerald2017-ci}, and inequitable exposures to social and environmental factors (e.g., environmental toxicants) can indirectly cause disparities in health outcomes \citep{Bellavia2018-environmental_mediators}. Additionally, disparities in access and trust in the healthcare system can lead to disparities in healthcare utilization \citep{Manuel2018-tk}. Differences in healthcare utilization can impact the completeness of observational health data in particular, as these data are only collected when individuals access healthcare services \citep{Sullivan2019-missing}. We argue that, when training and evaluating data-driven models for potential use in healthcare settings, these different "avenues" of bias must be characterized in a case-specific manner that accounts for known healthcare disparities in the specific disease domain. 

Many notions of fairness exist in machine learning \citep{gao_what_2025}. Group fairness is defined by splitting a population into groups based on some sensitive attribute(s) and enforcing equal model performance across subgroups according to some metric(s). However, notions of group fairness are limited by assumptions about the underlying data (e.g., demographic parity assumes equal rates of positive outcomes across all groups) and by the conflicting nature of group fairness metrics that can render simultaneous enforcement impossible \citep{Chouldechova2017-fairpred}. Individual fairness defines a similarity metric and seeks to ensure that individuals defined as "similar" by this metric have similar outcomes under the model \cite{dwork_fairness_2012}. However, individual fairness is susceptible to biased data and can overlook difficult-to-capture socioeconomic privileges and disparities in resource allocation \citep{Braveman2014-fk}. Causal fairness looks to identify and mitigate "unfair" causal pathways between some characteristic(s) and the outcome. Causal fairness approaches allow users to compare bias in the data to bias in the model, which is not possible when evaluating disparities based on subgroup comparison of model performance metrics. However, causal fairness can be limited by the need for domain knowledge to identify the underlying causal structure of the data and the need for statistical identifiability to estimate causal effects from observational data \citep{Pearl_00, Makhlouf2024-causality}. Additionally, calculating counterfactual quantities in high dimensions can be complex and unstable ~\citep{Mitra2022-causalityhighdim}. 

\begin{figure}[htbp]
\centering
\includegraphics[width=0.8\textwidth]{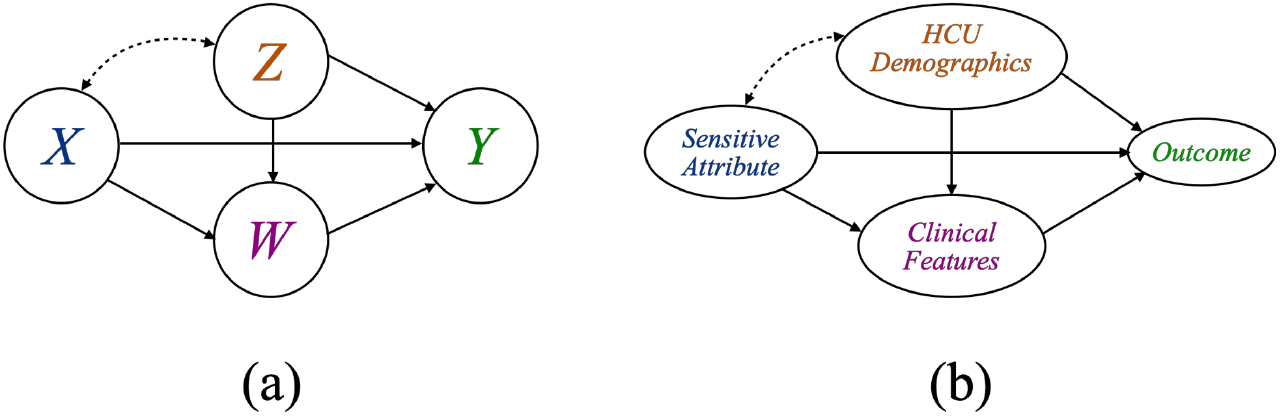}
\caption{Standard Fairness Model: (a) Overview of SFM as presented in \citet{plecko_cfatoolkit_2024} (b) Clinically motivated mapping of SFM for observational health data}
\Description{Causal directed acyclic graphs illustrating the standard fairness model and its mapping to observational health data, showing relationships between sensitive attributes, clinical features, healthcare utilization, and outcome}
\label{fig:sfm_dag}
\end{figure}

We argue that evaluation of fairness in healthcare should rely on \textit{path-specific causal fairness}, which defines biased causal paths in the data-generating mechanism and seeks to ensure that the model does not replicate or introduce bias through these "unfair" causal pathways. We focus on path-specific causal fairness over other methods, such as counterfactual fairness \cite{kusner_counterfactual_2017} because path-specific causal fairness allows us to decompose the biases in the healthcare system and form hypotheses about specific social and medical disparities that a model may be replicating. Our work builds on the structural fairness model (SFM) defined in \citet{plecko_cfatoolkit_2024}, which allows us to cluster variables into four groups (Figure \ref{fig:sfm_dag}a): sensitive attribute ($X$), outcome ($Y$), confounding variables ($Z$), and mediators ($W$). Through this clustering, we can ignore the relationships between variables within each cluster while maintaining identifiability of the relevant causal effects~\citep{Anand2023_clusterdags}. We offer a generalizable mapping of high-dimensional observational health data onto the SFM (Section \ref{sec:clinical_mapping}, Figure \ref{fig:sfm_dag}b), eliminating the need for domain knowledge of the case-specific causal system.

We establish a pipeline for enforcing and evaluating path-specific causal fairness in clinical risk prediction models. Users can select target effects for fairness interventions and explore a more multidimensional fairness-accuracy tradeoff that decomposes "bias" into specific paths that echo known health disparities. We demonstrate the pipeline's generalizability by applying it to four case studies. This generalizability is important given the rise of structured electronic health record (EHR) foundation models for healthcare \cite{pang2025fomohclinicallymeaningfulfoundation}, which require adaptable methods for models that may not be trained to explicitly address fairness. Three of our cases (acute myocardial infarction, systemic lupus erythematosus, type 2 diabetes mellitus) leverage embeddings from a structured EHR foundation model \cite{wornow2025contextcluesevaluatinglong} to train the baseline model on which fairness interventions are applied. Our fourth case (schizophrenia risk prediction) is built on a task-specific model to demonstrate the utility of our pipeline on a more "conventional" ML model. 

% In this work, we focus on the case study of schizophrenia risk prediction. Schizophrenia is a serious mental illness associated with increased rates of all-cause mortality and suicide \textcolor{red}{(cite)}, and early intervention through increased monitoring and/or psychiatric care can significantly improve clinical outcomes \textcolor{red}{(cite)}. Nevertheless, early diagnosis of schizophrenia remains a clinical challenge, as the disease shares psychosis-related symptomology with several other psychiatric disorders \textcolor{red}{(cite)}. Our model seeks to predict diagnostic transition from psychosis to schizophrenia. 

%Our work builds on the work from \cite{plecko_cfatoolkit_2024} and presents strategies to increase the utility of these methods in high-dimensional observational health data. 
Our contributions are as follows:
\begin{asparaenum}
    \item We offer a generalizable, clinically motivated mapping of tabular observational health data (e.g., EHR, administrative claims) onto the structural fairness model for framing clinical risk prediction problems. 
    \item We demonstrate the utility of dimensionality reduction methods to reduce the estimation error of doubly-robust path-specific causal effects. 
    \item We establish a pipeline for "diagnosing", enforcing, and evaluating path-specific causal fairness by contextualizing data-driven insights with known. healthcare disparities 
    \item We demonstrate that there are no universally applicable causal or non-causal fairness interventions that provide optimal accuracy-causal fairness tradeoffs and illustrate the importance of testing naive, feature selection-based fairness interventions alongside known causal regularization strategies. 
    \item We expand on prior work characterizing fairness-accuracy tradeoffs and present a more comprehensive set of tradeoffs between accuracy and \textit{path-specific} fairness, demonstrating the importance of context-specific prioritization of causal pathways. 
\end{asparaenum}
\section{Related Work}

\subsection{Enforcing causal fairness in machine learning}
Training causally fair models is an open challenge in machine learning. To train counterfactually fair models, \citet{kusner_counterfactual_2017} propose limiting a model to learn only from features that are non-descendants of the sensitive attribute; however, this technique is not feasible in observational health data, where most clinical variables could be causal descendants of the sensitive attribute ~\citep{gopal_implicithcbias, Meidert2023_biasreview, Gupta2023_intersectional}. Counterfactual fairness is also enforced using learned representations that are invariant or orthogonal to the sensitive attribute \cite{quinzan_learning_2022, chen_counterfactual_2025} or through counterfactual data augmentation  \citep{ma_learning_2023, zhou_counterfactual_2025, robertson_fairpfn_2024}. 
%  since access to healthcare is impacted by sensitive attributes in many healthcare settings ... Counterfactual fairness is distinct from path-specific notions that distinguish between different "paths" of unfairness (e.g., direct, indirect, and spurious effects). 

To enforce causal fairness with respect to path-specific measures,~\citet{plecko_cfatoolkit_2024} present an in-processing regularization to penalize causal effects along specific paths in the structural fairness model (SFM) and enforce fair decision-making ~\citep{plecko_outcomecontrol_2023}. Other work develops interventions to reduce only the direct effect \cite{distefano_directeffect_regularization_2020} or the effect of spurious features \citep{wang-etal-2021-enhancing}. Optimal transport has also been used to learn causally fair models \citep{de_lara_transport-based_2024, bayer_fair_2023, nabi_fair_2018}. In particular, \citet{nabi_learning_2019} propose to learn fair policies by sampling data from a `debiased' data-generating process. A full review of causality-based methods for fair ML can be found in \citet{su_review_2022}. We compare path-specific in-processing and causally fair resampling to naive feature selection methods for their potential to reduce path-specific effects and find that methods specifically designed to address causal fairness do not universally outperform naive methods, even when fairness is evaluated using path-specific causal effects. 
% ~\citet{kilbertus_avoiding_2017} presents a method to avoid discrimination via proxy variables. 

\subsection{Navigating the "fairness-accuracy tradeoff"}
Fairness and accuracy are important but potentially conflicting desiderata within ML \cite{gittens_fairnessrobustness}, with many works conceptualizing this tradeoff as a Pareto frontier  \cite{https://doi.org/10.1002/int.22354, 10.1145/3447548.3467326, xu_pareto_fairness}. Prior work in causality has aimed to measure this tradeoff, including by quantifying the extent to which model loss increases when trained under causal fairness constraints \cite{plecko_fairness-accuracy_2025} and by using the "average treatment effect" to assess the impact of fairness interventions \cite{10.1109/ASE56229.2023.00105}. However, the assumption that fairness and accuracy must exist as a "tradeoff" may be due to a lack of consideration of social and historical context \citep{10.1145/3447548.3467326}, limitations of the mathematical notations used to model fairness, and measurement error in biased data \citep{dutta_2020_fairacc}. Our work explicitly considers social context and baseline bias in the observational data by accounting for non-direct causal effects, as observational health data is known to reflect the biases of our medical system. For a more in-depth treatment of the necessity of the fairness-accuracy tradeoff, we refer readers to \cite{li_triangle_tradeoff_review}. We emphasize the importance of using path-specific causal fairness to assess ML models in the healthcare setting, breaking down bias into direct, indirect, and spurious causal effects. This allows us to consider the model in the context of broader health disparities, quantify the extent to which a learned model replicates and/or exacerbates unfair causal paths in the data, and measure the impact of fairness interventions on model performance.

\subsection{Fairness in foundation models}
Most work around fairness in foundation models focuses on language models \citep{Wang2025-llmfair_tutorial, Gallegos2024-llmfair_review} and vision models \citep{Ali2023-visionfomo_fair}. In medical imaging specifically, prior work found significant subgroup performance differences in segmentation foundation models \cite{Li_An_MICCAI2024, jin2024fairmedfm}. Other work echoes these findings and addresses these differences using balanced fine-tuning datasets \cite{pmlr-v225-khan23a} or "fair PCA post-processing" \cite{ali2023_cvfoundationmodels}. Consistent with broader ethical ML literature \cite{chen_biomedicalfairness_5things_2021}, \citet{queiroz2025fairfoundationmodelsmedical} discusses the ways that bias mitigation for medical vision foundation models must be an integrated effort involving investment at all steps in the AI pipeline, from data collection to training methodologies to deployment and regulation. Relatively little work focuses on fairness in tabular foundation models. \citet{robertson_fairpfn_2024} demonstrates how training on causally fair synthetic data can improve the fairness of tabular foundation models. In the case of in-context learning,  \citet{kenfack2025fairincontextlearningtabular} illustrates the promise of pre-processing fairness methods. Our work examines the path-specific causal fairness of structured (tabular) EHR foundation models, both at the pretraining step and after linear probing, which has received little attention in prior work. 
\section{Preliminaries}
\subsection{Standard Fairness Model}

%\textbf{Structural Causal Models (SCMs).}
%A Structural Causal Model (SCM)~\citep{Pearl_00} $\mathcal{M}$ is a quadruple $\mathcal{M} = \langle \mathbf{U},\mathbf{V}, P(\mathbf{U}), \mathcal{F} \rangle$, where $\mathbf{U}$ is a set of exogenous (latent) variables following a joint distribution $P(\mathbf{U})$, and $\mathbf{V}$ is a set of endogenous (observable) variables whose values are determined by functions $\mathcal{F} = \{f_{V_i}\}_{V_i  \in \mathbf{V}}$ such that $V_i \leftarrow f_{V_i}(\mathbf{pa}_{V_i}, \mathbf{u}_{V_i})$ where the parents  $\mathbf{Pa}_{V_i} \subseteq V$ and $\mathbf{U}_{V_i} \subseteq \mathbf{U}$. Each SCM $\mathcal{M}$ induces a distribution $P(\mathbf{V})$ and a causal graph $\mathcal{G} = \mathcal{G}(\mathcal{M})$ over $\mathbf{V}$ in which directed edges exist from every variable in $\mathbf{Pa}_{V_i}$ to $V_i$ and dashed-bidirected arrows encode common latent variables. An intervention is represented using the do-operator, $\operatorname{do}(\mathbf{X}=\mathbf{x})$, which encodes the operation of replacing the original equations of $X$ (i.e., $f_X(\mathbf{pa}_{X},\mathbf{u}_X)$) by the constant $\mathbf{x}$ for all $X \in \mathbf{X}$ and induces an interventional distribution $P(\mathbf{V} \mid \operatorname{do}(\mathbf{x)})$. For any $\mathbf{Y} \subseteq \mathbf{V}$, the \emph{potential response}  $\mathbf{Y}_{\mathbf{x}}(\mathbf{u})$ is defined as the solution of $\mathbf{Y}$ in the submodel $\mathcal{M}_{\mathbf{x}}$ given $\mathbf{U} = \mathbf{u}$, which induces a \emph{counterfactual} $\mathbf{Y}_{\mathbf{x}}$. 

The standard fairness model (SFM) \citep{plecko_cfatoolkit_2024} is a structural causal model (SCM) to assess causal fairness. An SCM \citep{Pearl_00} is a formalized model of a data-generating process that defines the causal relationship between each of the variables in the system. The SFM defines the sensitive attribute ($X$) that can directly impact the outcome ($Y$). Additionally, $X$ affects $Y$ through a set of mediator variables ($W$), confounded by $Z$ (Figure \ref{fig:sfm_dag}a). One key aspect of the SFM is the identifiability of the natural direct effect (NDE), natural indirect effect (NIE), and spurious effect (SE) from observational data. That is, under the assumption that the data-generating mechanism follows the SFM, all three effects can be estimated as nested counterfactuals from observational data (Appendix \ref{appendix:cfa_estimates}). 

\subsection{Clinical mapping of Standard Fairness Model}
\label{sec:clinical_mapping}
We leverage the SFM to define a generalizable procedure for mapping high-dimensional observational health data (e.g., EHRs, administrative claims) onto the SFM, thus providing a principled template for evaluating fairness in observational health settings. Such a generalizable mapping is important because relying on clinical domain knowledge to define structural assumptions is infeasible in high dimensions. Our mapping (Figure \ref{fig:sfm_dag}b) is as follows: as with the SFM, $X$ remains the sensitive attribute and $Y$ is the clinical outcome of interest. $W$ represents a high-dimensional representation of a person's clinical record that can include information derived from diagnoses, medications, procedures, devices, laboratory tests, etc. These features can be derived directly from a person's clinical record %(e.g., number of diagnoses, number of days prescribed a medication) 
or can be a learned representation of the clinical data presented as a set of embedding vectors. Finally, $Z$ represents other demographic variables, baseline comorbidities known before a visit, if available, and \textbf{healthcare utilization}. Healthcare utilization (HCU) refers to the frequency, types, and patterns of healthcare services that an individual uses. HCU can serve as a proxy for several unobserved variables such as social determinants of health (typically constituting latent common causes between $X$ and $Z$), including access to and trust in the healthcare system. In our work, we operationalize healthcare utilization as the frequency of visits. The NDE represents the "direct" effect of the sensitive attribute on the outcome; the NIE represents the effect of the sensitive attribute on the outcome as mediated by the rest of a person's observational health record ($W$). The SE represents the baseline change in the outcome distribution as a result of the two interventions $x=0$ and $x=1$ compared to their observational counterparts.
\section{Methods}

\begin{figure}[htbp]
\centering
\includegraphics[width=0.9\textwidth]{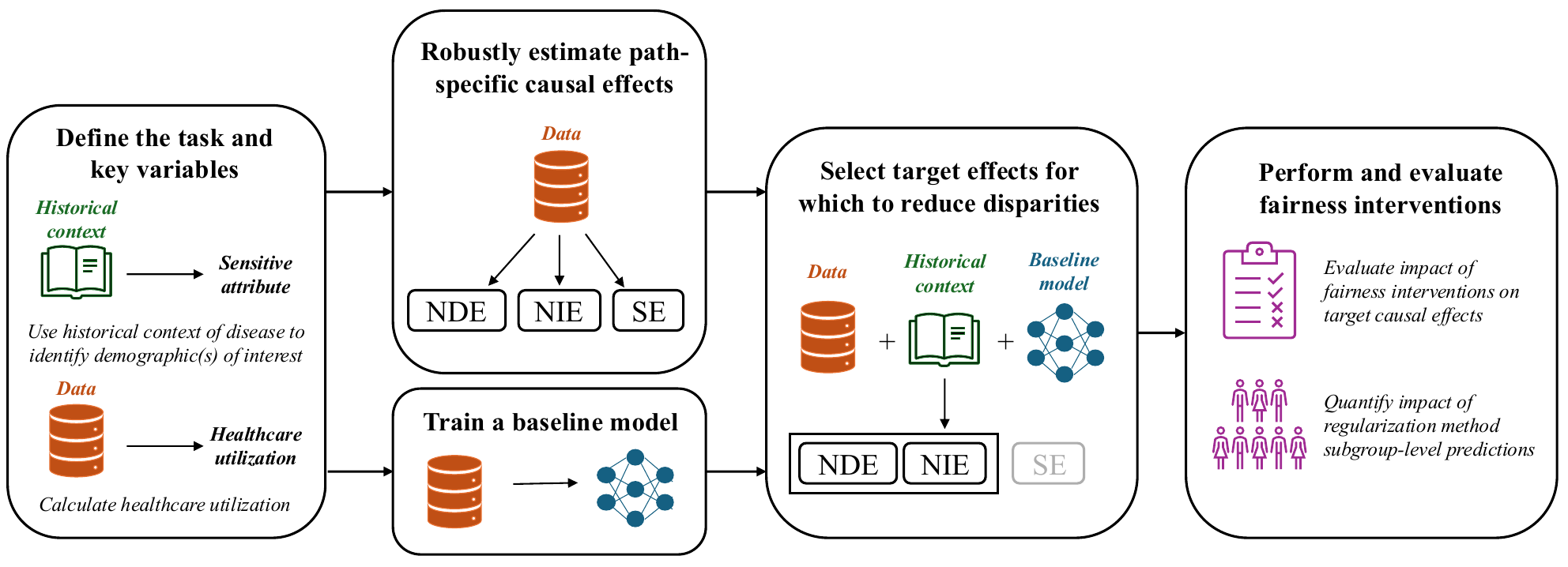}
\caption{Overview of the proposed workflow. Starting with task selection, we outline the steps required to train and evaluate a causally fair model. In particular, we find that data-driven decision making must be combined with known historical context about healthcare disparities specific to the clinical task at hand.}
\label{fig:workflow}
\Description{Flowchart presenting the workflow outlined in this paper. The steps are: 1) define the task and key variables (in particular, select the sensitive attribute and calculate healthcare utilization); 2) robustly estimate the path-specific causal effects; 3) train a baseline model; 4) select target effects (e.g., NDE, NIE, SE, some combination) for which to reduce disparities; 5) perform and evaluate causal fairness enhancing interventions.}
\end{figure}

In Figure \ref{fig:workflow}, we present a workflow for operationalizing the clinically motivated mapping of the SFM on a given problem that uses observational health data: i) define the task and key variables% (in particular, select the sensitive attribute and calculate healthcare utilization)
; ii) robustly estimate the path-specific causal effects; iii) train a baseline model; iv) select target effects (e.g., NDE, NIE, SE, some combination) for which to reduce disparities; v) perform and evaluate causal fairness enhancing interventions. 

% Operationalizing the causal fairness framework to train fair predictive models of the outcome $Y$ in healthcare requires the following considerations: i) capturing relevant but generalizable confounders $Z$, ii) knowledge of broader disparities characterized for the sensitive attribute and outcome of interest, iii) robustness of effect estimation in high dimensions to evaluate whether broadly known disparities are reproducible in the observational data at hand, iv) evaluating how a learned outcome model may exacerbate, reduce or replicate existing effects observed in the data, v) determining target regularization to mitigate specific effects (NDE, NIE, SE or a combination thereof), and evaluating the efficacy of the fairness intervention.

\subsection{Define the task and key variables}
Once a clinical prediction task is selected, we use this to set our outcome ($Y$) in the SFM. The sensitive attribute ($X$), however, still requires definition. $X$ is typically a discrete variable, often binary, for which we want to measure and enforce path-specific fairness. While purely data-driven analysis based on all known sensitive attributes is important, it is often more crucial and feasible to assess how broadly known disparities reflect in the data and downstream learned models. We advocate for selecting a sensitive attribute that is associated with known (or hypothesized) disparities in broader health literature. We use race or gender as our main sensitive attributes for each case; however, selecting subgroups for $X$ using an intersection of identities is possible under statistical feasibility (sufficient sample sizes to robustly evaluate path-specific effects). The next key variable(s), often ignored, are related to healthcare utilization. Healthcare utilization patterns drive predictive performance of learned models %, including foundation models
~\citep{zink2024accesscareimprovesehr, pang2025fomohclinicallymeaningfulfoundation} and serve as noisy proxies for unobserved social drivers of health. We define healthcare utilization as the number of visits per year. If there is consistent information about the visit setting (e.g., outpatient, inpatient, pharmacy), we create separate healthcare utilization features for each visit setting. These healthcare utilization features, in combination with any "non-sensitive" demographic features, comprise the confounder set of the SFM ($Z$). $Z$ must also include baseline health status, such as known comorbidities known at the beginning of the visit or stay, if data samples represent visits or stays. The clinical features ($W$) include all features used by the model that are not encapsulated by $X$, $Z$, and $Y$. 

\subsection{Robustly estimate path-specific causal effects}
\label{sec:methods_estimation}
%In this work, our goal is to estimate 1. the natural direct effect of the sensitive attribute ($X$) on the outcome ($Y$); 2. the natural indirect effect of the sensitive attribute ($X$) on the outcome ($Y$) as mediated by a high-dimensional set of clinical variables ($W$), and 3. the spurious effect of confounding variables (healthcare utilization and demographic variables, denoted by $Z$) on the relationship between $X$ and $Y$. 
% Given that the high-dimensional cluster of clinical features (W) are mediators in our structural causal model, propensity score-based strategies (e.g. LSPS, HDPS) for high-dimensional confounders are not applicable \textcolor{red}{cite; NOTE: not sure if this should go here or in related work}. 

Given the systemic nature of medical disparities and typical mechanisms these propagate, we argue that fairness should be evaluated using path-specific causal effects. These effects must be characterized in both the raw observational data and in learned models to assess the relative change in causal fairness.
%More specifically, these effects must be characterized in the raw observational data, and the relative change of NDE, NIE, and SE of a learned model must be assessed relative to the raw data effects and broadly known disparities. 
Steps iii) and iv), therefore, require robust estimation of the NDE, NIE, and SE from finite observational samples under current practice (raw data), a trained baseline model ($f: X, Z, W \mapsto Y$), and any trained model that includes a fairness intervention ~\citep{Jung2021-doublyrobust, plecko_cfatoolkit_2024}. %State-of-the-art work primarily relies on doubly robust estimators of these path-specific effects from finite samples in observational data~\citep{zhang2025pathspecific, plecko_cfatoolkit_2024}. 

Despite progress in doubly robust estimators, the dimensionality of $W$ and $Z$ remains a statistical bottleneck for finite sample estimators. We therefore assess the robustness of dimensionality reduction methods that provide the best tradeoff for robust path-specific estimation using synthetic data, where true effect sizes are known. We set the causal graph underlying the data-generating process to the SFM (Figure \ref{fig:sfm_dag}a) and initialize a different MLP to simulate each edge of the graph. We simulate synthetic data with dimensionality that closely matches the real data, with general function mappings (such as highly non-linear MLPs) to ensure that chosen dimensionality reduction will provide analogous robust estimates in real data where the ground-truth is unknown. %By separately simulating each edge, we can simulate $W$ and $Y$ "naturally" (randomly sampling the protected attribute $X$) and in the counterfactual setting forcibly setting $X$ to $0$ or $1$ and simulating $W_{X_{0}}$, $W_{X_{1}}$, $Y_{X_{0},W_{X_{0}}}$, $Y_{X_{0},W_{X_{1}}}$, $Y_{X_{0},W_{X_{1}}}$, and $Y_{X_{1},W_{X_{1}}}$. This allows us to calculate the "true" natural direct effect (NDE), natural indirect effect (NIE), and spurious effect (SE) as in \cite{plecko_cfatoolkit_2024}. 
Synthetic experiments show that doubly robust estimation alone still leads to large measurement errors in high dimensions ($\geq 500$ features) for the NDE and NIE (Figure \ref{fig:syntheticdata_nointervention}). We empirically evaluate methods of dimensionality reduction to identify the method(s) that lead to the most accurate path-specific effect estimates. Doubly robust estimators are effective because they only require one set of models (either the propensity models or outcome models) to be correctly specified~\cite{li_doubly_robust_2020}. Thus, we focus on dimensionality reduction methods that seek to retain predictive power about the sensitive attribute $X$ or $Y$. For each method listed below, we test reduction to $20\%$, $40\%$, $50\%$, $60\%$, and $80\%$ of the original dimensionality $|W|$. 

\begin{asparaitem}
\item \textbf{Learn $X$}: We retain predictive power about $X$ by identifying the subset of features in $W$ that demonstrate a distribution shift between $X=0$ and $X=1$ (i.e., select features with the largest shift in mean difference between the $X=0$ and $X=1$ subpopulations). 
\item \textbf{Learn $Y$}: We retain predictive power in $Y$ by identifying the "most important" subset of mediating features $W$ for predicting the outcome $Y$. We examine importance through LASSO logistic regression, commonly used for genetic mediation analysis \cite{jerolon_group_2024, ye_variable_2021}. Additionally, we train an XGBoost-based outcome model and identify features through %information gain (a feature importance metric specific to decision tree models) and 
permutation feature importance (PFI) \cite{fisher_pfi_2019}, a model-agnostic measure of feature importance. 
\item \textbf{Learn $W$}: While not directly supported by doubly robust theory, we test the impact of learning a condensed representation of $W$ that minimizes reconstruction loss. For this, we train a two-layer MLP autoencoder to generate a lower-dimensional embedding of $W$.
\end{asparaitem}

% We compare the following dimensionality reduction methods combined with doubly robust estimation: 
% \begin{itemize}
% \item \textbf{Identifying the subset of mediating features $W$ that are most predictive of the outcome $Y$:} We examine feature importance through LASSO logistic regression, a method commonly used for genetic mediation analysis \cite{jerolon_group_2024, ye_variable_2021}. Additionally, we train an XGBoost model to learn $Y$ from $W$ and measure feature importance based on information gain (a feature importance metric specific to decision tree models) and permutation feature analysis \cite{fisher_pfi_2019}, a model-agnostic measure of feature importance. 
% \item \textbf{Identifying the subset of features in $W$ that demonstrate a distribution shift between $X=0$ and $X=1$:} We select features with the largest shift in mean difference between the $X=0$ and $X=1$ subpopulations. 
% \item \textbf{Creating a faithful low-dimensional embedding of $W$:} We train a two-layer MLP autoencoder to generate a lower-dimensional embedding of $W$. This model is trained using reconstruction loss (mean squared error). 
% \item \textbf{Creating a low-dimensional embedding of $W$ that retains information about  $X$ and $Y$:} We train the autoencoder described above to generate a lower-dimensional embedding of $W$; however, we combine three losses: reconstruction loss of $W$, binary cross entropy with respect to $X$, and binary cross entropy with respect to $Y$. 
% \end{itemize}

\subsection{Train a baseline model}
The baseline model is trained on all variables in $X$, $W$, and $Z$. Our pipeline is model-agnostic, so any model architecture can be used. The evaluation of causal effects on model output requires "hard" (binary) labels; we create these labels by selecting the threshold probability that jointly optimizes sensitivity and specificity in the validation dataset. 

\subsection{Select target effects for which to reduce disparities}
\label{sec:methods_selectingeffects}
Once we select a clinical problem and establish a baseline model, we should select the causal paths to prioritize for elimination. This idea of prioritizing causal pathways is similar to the notion of "business necessity" \cite{plecko_cfatoolkit_2024}; however, rather than ignoring bias through specific pathways, we recognize that regularization across all causal pathways may not be feasible while maintaining model performance ("no free lunch"). Thus, we advocate for case-specific prioritization of causal pathways that accounts for domain knowledge of health disparities and relative disparities in the data compared to baseline models. We examine 1) the relative sizes of each causal effect in the raw data and the ways in each observed causal effect may be reflective of known healthcare disparities in the specific clinical domain; 2) how causal effects may be exacerbated in the baseline model compared to the raw data; and 3) the contextual factors (either from health disparities literature or algorithmic fairness literature) that may explain the causal effects of the baseline model.

\subsection{Perform and evaluate fairness interventions}\label{sec:fairness_interventions}
Causal regularization is often proposed as an effective method to mitigate path-specific unfairness due to its ability to target individual or a combination of effects. Feature selection is a commonly used strategy for training model fairness over a range of fairness definitions \citep{Belitz2021-cw, Galhotra2022-fw, Yang2023-gm, Njoku2025-mf}. Complementarily, non-causal regularization techniques that operationalize notions of group and individual fairness are prominent in algorithmic fairness literature. We evaluate the following \textbf{causal regularization} strategies: 
\begin{asparaitem}
    \item \textbf{Path-specific inprocessing}: We implement the path-specific inprocessing approach from \citet{plecko_cfatoolkit_2024}. This approach leverages the model output for the true data and the counterfactual (changing a person's race or gender) to estimate the NDE, NIE, and SE. It then penalizes these effects in the loss function. We train models that regularize a single effect (NDE-only, NIE-only, SE-only) and models that regularize with respect to all three effects. %In our experiments, we vary the regularization strength parameter between 0.1 and 100 and select the model with the best validation-set tradeoff between BCE loss and average causal effect. 
    \item \textbf{Causally fair resampling}: We implement a version of the approach from \citet{nabi_learning_2019}, which blocks unfair causal pathways by learning fair distributions over the sensitive attribute and mediator. The method computes path-specific effects using inverse probability weighting and employs constrained optimization to find model parameters that minimize the targeted path-specific effect. We average over the sensitive attribute and mediators using the learned fair distributions, thus removing the influence of the sensitive attribute through the targeted causal pathways. %In our experiments, we apply this approach to target both the NDE and the total effect of the sensitive variable on the outcome. As our mediator is high-dimensional, we apply Bayes' rule to compute the density ratio required for estimating the NDE by modeling the reverse conditional distribution. We derive the fair distribution over the sensitive variable by averaging over the observed training data, thus approximating the theoretical integral over all possible mediator values, and use this to obtain fair predictions.
\end{asparaitem}

We evaluate multiple \textbf{feature selection}-based strategies: 
\begin{asparaitem}
    \item \textbf{Demographic unawareness}: We train models without any explicit demographic information (race and gender), including the sensitive attribute. This experiment is representative of the idea of "fairness through unawareness", a simple but widely debated debiasing method in fair ML and ML for health \cite{holtgen_fairnessunawareness, basu_use_2023}.
    \item \textbf{"Unbiased" feature selection}: We identify "highly biased features" as those with a large standardized mean difference (SMD) between the specified sensitive attribute and remove these features during model training. For a specified feature count ($n$), we keep the $n$ features with the lowest SMD. %We train models with varying feature counts and select the model with the best validation-set tradeoff between binary cross entropy (BCE) loss and average causal effect (NDE + NIE + SE). 
    \item \textbf{Greedy feature selection} We greedily select features that maximize predictive performance while minimizing bias. We quantify a feature's contribution to predictive performance using a feature importance metric and quantify its contribution to bias using SMD, as defined above. The relative contribution of SMD to the feature score is selected based on the "elbow point" that maximizes distance from the baseline importance-SMD tradeoff. %As described above, we vary feature count and select the number of features based on BCE loss and average causal effect. 
\end{asparaitem}

Finally, we test the following \textbf{non-causal regularization} strategies:  
\begin{asparaitem}
    \item \textbf{Learning fair representations}: We investigate representations learned to optimize individual fairness model proposed in~\citet{pmlr-v28-zemel13}. This is a pre-processing method that seeks to learn a representation of the data that obscures information about the sensitive attribute while maintaining predictive power. 
    \item \textbf{Equalized Odds}: We investigate the impact of imposing a group fairness constraint on path-specific causal fairness. Specifically, we operationalize equalized odds \cite{hardt_eqodds} by adding a regularization term to the loss function that penalizes differences in the true and false positive rates. %In our experiments, we vary the regularization strength parameter between 0.1 and 100 and select the model with the best validation-set tradeoff between BCE loss and average causal effect. 
\end{asparaitem}

We evaluate the models based on performance (AUROC) and path-specific causal effects (focusing on the "prioritized" causal pathways, as described in Section \ref{sec:methods_selectingeffects}). In order to better characterize the impact of the regularized prediction model on specific subgroups of patients, we also report the Pearson Correlation between the sensitive attribute and the true outcome, as well as between the sensitive attribute and the predicted outcomes (Appendix~\ref{appendix:ami_additional_results}-\ref{appendix:scz_additional_results}). This allows us to compare the relationship between the sensitive attribute and the outcome in the data, baseline model, and the "fair" model. This analysis is especially important for tasks where our goal is to improve racial fairness, as we report the causal effects with respect to Black vs. White patients, and assessing the impact on other racial groups is crucial. 
\section{Experiments and Results}
We demonstrate our proposed workflow on four clinical risk prediction tasks: acute myocardial infarction (AMI),  systemic lupus erythematosus (SLE), type 2 diabetes mellitus (T2DM), and schizophrenia (SCZ). We select these tasks because they span a diverse set of clinical domains and are all associated with known health disparities (Section \ref{sec:task_definition}). In each case, we select an "at-risk" cohort of individuals with related symptoms or comorbidities of the illness, and seek to predict which of these individuals will go on to develop the given disease. Our framing of tasks within the "at-risk" cohort rather than a general population increases the task difficulty because many well-established risk factors for the disease are widely present in the data and are therefore less predictive of the outcome of interest. This forces the model to rely less on risk factors that clinicians are already attuned to, thus increasing potential for clinical utility. For AMI, SLE, and T2DM, we use the phenotypes defined in a prior foundation model benchmark \citet{pang2025fomohclinicallymeaningfulfoundation} and retrain the LLAMA foundation model on CUMC-EHR. We select the LLAMA model because is a widely used architecture used both within and outside the healthcare setting \cite{wornow2025contextcluesevaluatinglong}.

\subsection{Define the task and key variables}
\label{sec:task_definition}
We demonstrate our pipeline in two datasets representing typical structured observational health data sources: CUMC-EHR, a deidentified EHR of $\sim$5.3 million patients from a large urban academic medical center, and the deidentified Merative Multi-State Medicaid dataset (MDCD). The MDCD dataset includes all data collected from healthcare providers for medical billing for $25$ million patients from $11$ different US states. Both datasets are structured using the Observational Medical Outcomes Partnership Common Data Model \cite{Hripcsak2019-ks}, ensuring reproducibility in other compatible health datasets. In both datasets, gender is recorded as male or female and the method of capture for this variable (e.g., patient-reported, assigned at birth) is not recorded and may vary between patients. We are therefore unable to extend our analyses beyond binary gender. In CUMC-EHR, race is captured as a categorical variable: American Indian or Alaskan Native, Asian, Black, Native Hawaiian or Pacific Islander, White, Missing, and Other. In MDCD, race is captured with three categories: Black, White, and Missing. The method of capture for race is also unknown in both datasets. 

\subsubsection{Acute Myocardial Infarction (AMI)}
We predict AMI in patients with prior ischemic heart disease. We focus on mitigating gender disparities ($x_{0} = female, x_{1}=male$) due to known gender-based differences in presenting symptoms that can lead to underdiagnosis and delayed diagnosis of AMI in women~\citep{Kim2023-cz, Ngaruiya2024-vc, Imboden2026-ri}. We predict AMI onset using the embeddings from a LLAMA foundation model \cite{wornow2025contextcluesevaluatinglong} pre-trained on CUMC-EHR; the model uses RoPE embeddings to encode temporal information and is trained using next token prediction (NTP) with context length $8192$. The model embeddings represent a person's clinical history ($W$) and can be optionally combined with $X$ and $Z$ to predict the downstream outcome. This dataset does not contain comprehensive information about visit settings; thus, we represent healthcare utilization as a single variable defined as the number of distinct days with clinical "touchpoints" (visits, diagnoses, or procedures) per year. $Z$ consists of healthcare utilization and patient race. 

\subsubsection{Systemic Lupus Erythematosus (SLE)}
We predict the first occurrence of SLE among individuals with at least one SLE-related symptom or a prescription of a drug used to treat SLE symptoms. We focus on mitigating gender disparities ($x_{0} = female, x_{1}=male$) in the case of SLE due to differing diagnosis rates ($4-22\%$ of diagnosed SLE patients are male) and significantly differing symptom profiles across sexes \cite{Tan2012-jv, Do_Socorro_Teixeira_Moreira_Almeida2011-lw}. Additionally, prior experimental work has found that clinicians faced with otherwise identical symptom profiles are less likely to diagnose men with SLE \cite{Simard2022-zz}. We use the same data source (CUMC-EHR), LLAMA model, and healthcare utilization as outlined above for AMI. 

\subsubsection{Type 2 Diabetes Mellitus (T2DM)}
We predict T2DM onset from the at-risk cohort defined in \citet{pang2025fomohclinicallymeaningfulfoundation} and a validated diabetes phenotype definition \citep{Suchard2021LEGENDT2DM}.  We focus on mitigating racial disparities ($x_{0} = White, x_{1}= Black$). T2DM is known to have a higher prevalence among Black individuals compared to White individuals, likely due to disparities in socioeconomic status that increase diabetes risk %(e.g., limited access to healthy food, health education, and increased environmental stressors) 
\citep{Deng2025-bj, Hackl2024-ip}. In addition to racial differences in prevalence, we find that several rule-based clinical algorithms for quantifying diabetes risk appear to underestimate T2DM risk for Black individuals, likely due to the absence of social and economic determinants of health from these algorithms~\citep{Cronje2023-vv}. It is important to measure the extent to which ML-based clinical prediction algorithms may remedy or exacerbate the biases found in such non-ML clinical algorithms. We use CUMC-EHR, the LLAMA model, and the healthcare utilization calculation for task definition as outlined above. However, since our sensitive attribute is patient race, we include patient gender in $Z$. 

\subsubsection{Schizophrenia (SCZ)}
For the case of SCZ risk prediction, our "at-risk" cohort is defined as individuals with at least 3 years of observation and at least one psychosis diagnosis, and we predict schizophrenia onset one year after the first psychosis visit. We mandated three years of observation because external validation of the SCZ phenotype demonstrates that requiring three years of prior observation improves the positive predictive value from $69\%$ to $77\%$ \citep{finnerty_prevalence_2024}. Black individuals are overdiagnosed with schizophrenia due to differences in the interpretation of affective symptoms \cite{trierweiler_clinician_2000, Gara2019-bp}. Additionally, rates of SCZ diagnosis following psychosis are higher for Black individuals compared to white individuals \cite{Moe2024-hv}. For these reasons, we focus on mitigating racial disparities ($x_{0} = White, x_{1} = Black$). The SCZ prediction model is trained for this single task using the MDCD data. We generated features ($W$) based on frequency of conditions, laboratory tests, procedures and duration of prescriptions. We define healthcare utilization through $10$ setting-specific frequency variables (e.g., outpatient, inpatient). $Z$ consists of healthcare utilization and patient gender. 
% In the task-specific schizophrenia prediction model, the NDE represents the difference in the schizophrenia diagnosis probability ($Y$) if an individual observed a Black patient and a White patient who both had clinical histories that reflect how a White individual is treated in the medical system. The NIE represents the difference in the schizophrenia diagnosis probability for a Black individual if they were to have a clinical history drawn from a distribution of white patients versus drawn from a distribution of Black patients. 

\begin{figure}[htbp]
\centering
\includegraphics[width=1\textwidth]{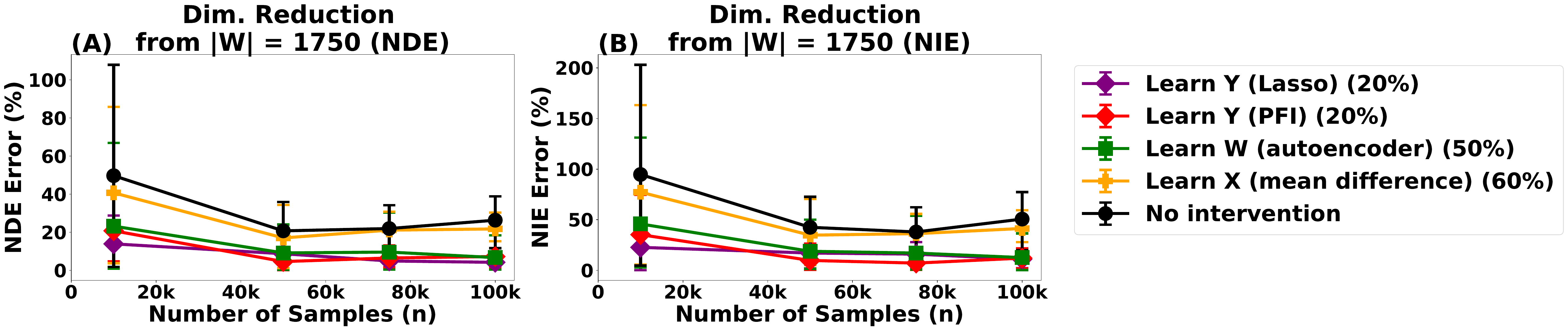}
\caption{Error in the path-specific causal estimates for dimensionality $|W|=1,750$. Each line corresponds to a different dimensionality reduction method, with the black line corresponding to no dimensionality reduction. The NDE estimation error is reflected in part (a), and the NIE estimation error is reflected in part (b). We find that interventions that identify features important for predicting the outcome ($Y$) outperform all other methods.}
\label{fig:syntheticdata_withintervention_1750}
\Description{Side-by-side line plots with associated error bars. Each line is a different color, corresponding to the given legend of dimensionality reudction techniques}
\end{figure}

\subsection{Robustly estimate path-specific causal effects}
\label{sec:synthetic_estimation_results}
Using synthetic data, we demonstrate that estimating causal effects in high mediator dimensions $(|W| \geq 500)$ leads to large estimation errors in the NDE and NIE %This error is most pronounced in the case of small samples ($n = 10,000$), but even in larger samples ($n = 100,000$), the NDE and NIE error can exceed $20\%$ and $30\%$, respectively 
(Figure~\ref{fig:syntheticdata_nointervention}, Table \ref{tab:estimation_error_pfi}). We test each of the dimensionality reduction methods outlined in Section \ref{sec:methods_estimation} on various mediator dimension sizes, ranging from $|W| = 750$ to $|W| = 1,750$. We consistently find that dimensionality reduction methods that identify the most important features for predicting $Y$, particularly PFI, are most effective at reducing both the NDE and NIE estimation error across all sample sizes (Figure \ref{fig:syntheticdata_withintervention_1750}). The spurious effect calculation is not impacted by changing the representation of $W$. These results are broadly consistent for $|W|=750$, $|W|=1,000$ and $|W|=1,500$ (Appendix \ref{appendix:estimation}). In the case of AMI, SLE, and T2DM, we learn the features most predictive of $Y$ through PFI \citep{fisher_pfi_2019}, as this yields the best results in the experiments where $|W|=750$ (and the clinical embeddings are all $|W|=768$). In the case of SCZ, we select highly predictive features using augmented feature occlusion (AFO), a permutation analysis method appropriate for temporal models \citep{tonekaboni_what_2020}.

\subsection{Train a baseline model}

For the AMI, SLE, and T2DM prediction tasks, we pre-train a LLAMA model to generate embeddings (embedding dimension $=768$) that represent a person's clinical history. We use these embeddings, demographic features, and healthcare utilization information to train a logistic regression model as the baseline for each of these three tasks. We focus this approach rather than on finetuning the full foundation model due to its lower computational cost and relatively high performance \citep{pang2025fomohclinicallymeaningfulfoundation}. For the SCZ prediction task, we employ a task-specific model: we bin data into 90-day increments and train an encoder-only transformer model using BCE loss. 
% SJ is here.
\begin{table}[ht]
\centering
\resizebox{\textwidth}{!}{%
    \begin{tabular}{lcccccc}
    \toprule
    & \multicolumn{2}{c}{\textbf{NDE}} & \multicolumn{2}{c}{\textbf{NIE}} & \multicolumn{2}{c}{\textbf{SE}} \\
    \cmidrule(lr){2-3} \cmidrule(lr){4-5} \cmidrule(lr){6-7}
    \textbf{Disease} & Data & Baseline model & Data & Baseline model & Data & Baseline model \\
    \midrule
    
\textbf{AMI} & 0.003 (-0.003, 0.004) & 0.073 (0.032, 0.077) & 0.008 (0.005, 0.009) & 0.153 (0.143, 0.157) & -0.001 (-0.001, -0.000) & -0.005 (-0.008, -0.004) \\
\textbf{SLE} & -0.002 (-0.002, -0.001) & -0.038 (-0.047, -0.025) & 0.000 (-0.000, 0.000) & -0.012 (-0.021, -0.008) & 0.000 (-0.000, 0.000) & -0.002 (-0.002, -0.001) \\
\textbf{T2DM} & 0.014 (0.002, 0.016) & 0.033 (0.027, 0.050) & 0.005 (-0.004, 0.007) & 0.059 (0.053, 0.107) & 0.002 (0.000, 0.002) & 0.003 (-0.015, 0.007) \\
\textbf{SCZ} & 0.018 (0.005, 0.031) & 0.120 (0.104, 0.134) & 0.017 (0.003, 0.027) & 0.015 (-0.000, 0.039) & -0.016 (-0.017, -0.012) & -0.069 (-0.073, -0.061) \\
\bottomrule
\end{tabular}%
}
\caption{We report the Natural Direct Effect (NDE), Natural Indirect Effect (NIE), and Spurious Effect (SE) in the dataset and the baseline model for each prediction task. We report the effect size based on the held-out test set to remain consistent with the evaluation of model regularization methods and report accompanying $95\%$ confidence intervals derived through bootstrapping.}
\label{table:rawdata_effects}
\end{table}

\subsection{Select target effects for which to reduce disparities}
For each task, we report the path-specific causal effects for the raw data and baseline models in Table \ref{table:rawdata_effects} and contextualize this information with known healthcare disparities to select which causal effects we aim to focus on. 

\subsubsection{Acute Myocardial Infarction (AMI)}
When examining causal effects in the data, only the NIE is significant ($0.008$ [$95\%$ CI: $0.005, 0.009$]); the positive value of the NIE indicates that an individual with clinical features drawn from a "male" population distribution is more likely to be diagnosed with AMI. This is consistent with findings that clinicians underdiagnose women with AMI due to a differing "typical" symptom profile \citep{Kim2023-cz, Ngaruiya2024-vc, Imboden2026-ri}. In the baseline model, both the NDE and NIE increase by an order of magnitude (NDE: $0.073$; NIE: $0.153$), while the SE remains relatively small ($-0.005\ [-0.008,\ -0.004]$). The increase in NDE and NIE indicates that the model introduces direct bias against women that does not exist in the dataset, and that the model is more tuned to a "classically male" symptom profile for AMI compared to clinical practice. We therefore focus on jointly reducing the NDE and NIE for AMI risk prediction. 

\subsubsection{Systemic Lupus Erythematosus (SLE)}
When examining the impact of gender on SLE, the NIE and SE are insignificant, and the NDE is small ($-0.002\ [-0.002, -0.001]$). The NDE and NIE increase significantly in the baseline model (NDE: $-0.038\ [-0.047,\ -0.025]$; NIE: $-0.012\ [-0.021,\ -0.008]$), with the NDE increasing more than the NIE. The negative NDE indicates that women are more likely to be diagnosed than men, consistent with the low diagnosis rates of SLE in men \citep{Tan2012-jv, Simard2022-zz}. The NIE suggests that individuals with a "male" distribution are more likely to be diagnosed with SLE than individuals with data drawn from a "female" distribution. Due to direct bias against men when diagnosing SLE and the larger size of the NDE in both the data and the baseline model, we focus on reducing the NDE for SLE. 

\subsubsection{Type 2 Diabetes Mellitus (T2DM)}
In the T2DM cohort, the dataset reflects a significant NDE ($0.014\ [0.002, 0.016]$) that indicates overdiagnosis of T2DM in Black individuals, consistent with epidemiological literature \citep{Deng2025-bj, Hackl2024-ip}. The NDE is exacerbated in the baseline model ($0.033\ [0.027, 0.134]$), which also worsens the NIE ($0.059\ [0.053, 0.107]$). The NIE suggests that an individual with clinical features drawn from a "white" population distribution is more likely to be diagnosed with T2DM; this is consistent with known algorithmic biases \citep{Cronje2023-vv}. The SE remains insignificant in the data and baseline model. Thus, we focus on jointly reducing the NDE and NIE for T2DM risk prediction.  

\subsubsection{Schizophrenia (SCZ)}
The NDE, NIE, and SE are all similar in magnitude (NDE: $0.018\ [0.005,\ 0.031]$; NIE: $0.017\ [0.003, 0.027]$; SE: $-0.016\ [-0.017,\ -0.012]$) in the data. The positive NDE is reflective of overdiagnosis of SCZ in Black patients compared to white patients, while the positive NIE is reflective of the fact that an individual with clinical features drawn from a "white" population distribution is more likely to be diagnosed with SCZ. The negative SE suggests that the confounder variables in $Z$ (gender, healthcare utilization) may play a larger role in diagnosis for Black patients than white patients. Of these three effects, the NDE reflects known overdiagnosis of SCZ in Black individuals \citep{trierweiler_clinician_2000}, while the NIE and SE may reflect broader disparities in quality of care \citep{Fiscella2016-tf} and healthcare utilization \cite{Manuel2018-tk}, respectively. In the baseline model, we observe a significant exacerbation in NDE (Model: $0.120\ [0.104,\ 0.134]$) and SE (Model: $-0.069\ [-0.073, -0.061]$). The NIE remains similar the data ($0.015\ [0,\ 0.039]$). These causal effects suggest that the model is more reliant on race than a clinician might be (NDE), and that the disparate weight on confounding variables (healthcare utilization, gender) is more pronounced in the model than in clinical practice (SE). %It is also important to note that healthcare utilization is a potential predictor of schizophrenia onset, as individuals with schizophrenia demonstrate higher utilization even when compared to individuals with other serious mental illnesses \citep{Wallace2020-ih, Jonsson2024-bj}. 
Based on the tenfold increase in NDE between the raw data and the baseline model and the known problems with overdiagnosis of schizophrenia in Black individuals, we prioritize reducing the NDE in the SCZ risk prediction case. Additionally, we treat reduction of SE as a secondary target, as the effect may be due to disparities in healthcare utilization.

% The negative SE indicates that the impact of severing the bidirectional causal link between $X$ (race) and $Z$ (gender, healthcare utilization) has a larger impact for Black individuals than for white individuals. For Black patients, moving from the observational distribution (i.e. looking at the schizophrenia diagnosis probability for individuals who happen to be Black) to the experimental distribution (i.e. probability of schizophrenia diagnosis if we simulate that everyone's race is Black) increases the expected probability of diagnosis. For white patients, moving from the corresponding observational distribution to the experimental distribution slightly decreases the expected probability of diagnosis. This suggests that the confounder variables in $Z$ may play a larger role in diagnosis for Black patients than white patients. 

% Overall, the magnitudes of all three causal effects are similar; this suggests that none of these effects dominate the landscape of bias. Rather, there is a complex balance of direct bias, indirect bias mediated through clinical features, and spurious bias (introduced via healthcare utilization and other demographic variables) that are all relevant to understanding the health disparities present in schizophrenia diagnoses. 

\subsection{Perform and evaluate fairness interventions}
\label{sec:fairness_intervention_results}
\begin{figure}[htbp]
\centering
\includegraphics[width=1\textwidth]{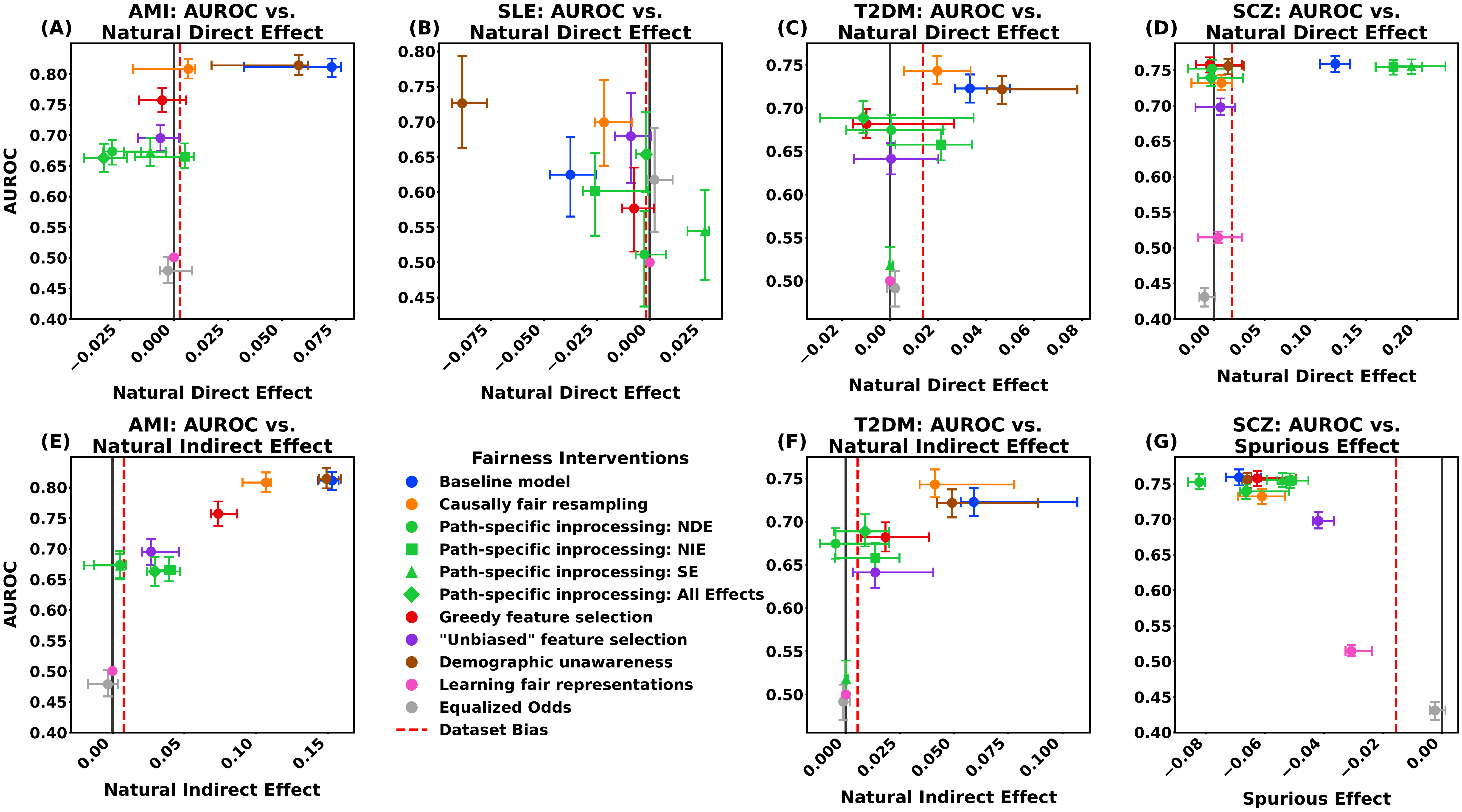}
\caption{Tradeoffs between model performance and target path-specific effects across all tasks. We plot the target path-specific effects for each intervention (NDE, NIE, and/or SE) against the model performance (AUROC) for each of the "target" effects (Section \ref{sec:methods_selectingeffects}). (A) and (E) show the NDE and NIE for AMI; (B) shows the NDE for SLE; (C) and (F) show the NDE and NIE for T2DM; (D) and (G) show the NDE and SE for SCZ.}
\label{fig:all_tradeoffs_plot}
\Description{4 columns by 2 rows of scatter plots. Points are colored differently based on the fairness intervention (included in the legend) and have associated error bars. The x-axis is the path-specific causal effect and the y-axis shows AUROC.}
\end{figure}

We discuss the efficacy and impact of all interventions outlined in Section~\ref{sec:fairness_interventions} for each clinical prediction task. We present the AUROC-effect tradeoff for each of the "target" effects (Figure \ref{fig:all_tradeoffs_plot}).

\subsubsection{Acute Myocardial Infarction (AMI)}
As shown in Figure \ref{fig:all_tradeoffs_plot}A and \ref{fig:all_tradeoffs_plot}E, the baseline model and demographic-unaware model are associated with the highest model performance (Baseline AUROC: $0.812\ [0.796,\ 0.825]$; demographic unawareness: $0.814\ [0.799,\ 0.832]$). However, both models have large NDE ($>0.05$) and NIE ($>0.14$). Causally fair resampling maintains near-equal performance to the baseline model ($0.809\ [0.794,\ 0.825]$) while eliminating the NDE; however, the NIE is still relatively large ($-0.111\ [0.095,\ 0.115]$). Greedy feature selection also eliminates the NDE and improves the NIE beyond the baseline and causally fair resampling ($0.074\ [0.069,\ 0.087]$); however, there is a $6.8\%$ drop in AUROC ($0.757\ [0.738,\ 0.777]$). If we allow model performance to drop even further, the "unbiased" feature selection intervention (AUROC: $0.695\ [0.674,\ 0.716]$) reduces the NIE even further ($0.027\ [0.021,\ 0.046]$) while eliminating the NDE. Through this, we observe that it is possible to eliminate the NDE without sacrificing model performance, but there is a clear tradeoff between reducing the NIE and maintaining model performance (Appendix \ref{appendix:ami_additional_results}). %We find that all three fairness interventions significantly reduce the correlation between gender and predicted model output relative to the baseline model (\textcolor{red}{Figure TODO}, Appendix \ref{appendix:ami_additional_results}). 

%\begin{figure}[htbp]
%\centering
%\includegraphics[width=1\textwidth]{figures/all_diseases_correlations.pdf}
%\caption{Correlation between demographics and (predicted) outcome. We measure the correlation between the demographic and the outcome in the data, as well as the demographic and the predicted outcome in the baseline model and the top fairness interventions for the relevant demographic group and fairness interventions.}
%\label{fig:all_correlations_plot}
%\end{figure}

\subsubsection{Systemic Lupus Erythematosus (SLE)}
As shown in Figure \ref{fig:all_tradeoffs_plot}B, the demographic-unaware model is associated with the best AUROC ($0.726\ [0.662,\ 0.794]$). However, the NDE ($-0.089\ [-0.094,\ -0.077]$) is significantly higher compared to the data and the baseline model. The next-best model performances (outperforming even the baseline model) are associated with causally fair resampling ($0.699\ [0.638,\ 0.759]$) and selection of "unbiased" features ($0.680\ [0.613, 0.741]$). The causally fair resampling reduces, but does not eliminate the NDE ($-0.030\ [-0.044,\ -0.030]$) and "unbiased" feature selection eliminates the NDE (Appendix \ref{appendix:sle_additional_results}). %We find that both fairness interventions significantly reduce the correlation between gender and predicted model output relative to the baseline (demographic unaware) model, with causally fair resampling rendering this correlation equal to that in the data (\textcolor{red}{Figure TODO}, Appendix \ref{appendix:sle_additional_results}).

\subsubsection{Type 2 Diabetes Mellitus (T2DM)}
The highest performing model that eliminates both the NDE (Figure \ref{fig:all_tradeoffs_plot}C) and NIE (Figure \ref{fig:all_tradeoffs_plot}F) is the one trained using path-specific in-processing. The AUROC of this model ($0.689\ [0.671, 0.709]$) represents a $7.3\%$ drop in performance relative to the highest-performing model, which uses the causally fair resampling intervention (AUROC: $0.743\ [0.728, 0.760]$). However, causally fair resampling maintains a significant NDE ($0.020\ [0.006,\ 0.034]$) or the NIE ($0.041\ [0.034,\ 0.078]$). Thus, we observe a tradeoff between model performance and model bias along both the NDE and NIE (Appendix \ref{appendix:t2dm_additional_results}). %We compare the correlation between all available patient race groups and (predicted) outcome (Figure \textcolor{red}{todo}). We find that, relative to the best-performing model (causally fair resampling), the "fair" model (path-specific inprocessing) demonstrates lower magnitude of correlation between patient race and predicted output for White individuals but that both models exacerbate this correlation relative to the data for individuals noted as "missing/other" race. Correlations within the data and both models are similar for Asian, American Indian/Alaskan Native, and Native Hawaiian/Pacific Islander individuals. Additional discussion of these correlations can be found in Appendix \ref{appendix:t2dm_additional_results}

\subsubsection{Schizophrenia (SCZ)}
As shown in Figure \ref{fig:all_tradeoffs_plot}D and \ref{fig:all_tradeoffs_plot}G, the baseline model is associated with the best performance (AUROC: $0.759\ [0.748,\ 0.770]$) but larger NDE and SE compared to the data (NDE: $0.120\ [0.104,\ 0.134]$; SE: $-0.069\ [-0.073,\ -0.061]$). Greedy feature selection maintains the performance of the baseline model (AUROC: $0.758\ [0.747,\ 0.768]$) while eliminating the NDE and maintaining a similar SE to the baseline model ($-0.063\ [-0.065,\ 0.049]$). If we allow for deterioration in performance, we find that "unbiased" feature selection (AUROC: $0.698\ [0.687,\ 0.710]$) eliminates the NDE and reduces the SE to $-0.042\ [-0.044, -0.037]$. Thus, while we do not observe a tradeoff between performance and NDE, we do observe a tradeoff between performance and SE (Appendix \ref{appendix:scz_additional_results}). %When we examine the correlation between patient race and (predicted or actual) output, we find again that the "fair" models lower the magnitude of correlation relative to the baseline model. All correlations for individuals with "missing" race are small ($<0.015$) (Figure \textcolor{red}{todo}, Appendix \ref{appendix:scz_additional_results}). 

\section{Discussion}
Our pipeline enables the following insights on enabling fairness-enhancing interventions in observational health data. 

\textbf{Mapping observational health data to the SFM enables generalizable evaluation of causal fairness.} Our mapping can be used for multiple clinical risk prediction problems and works with clinical variables ($W$) generated directly from the data and with feature embeddings generated through foundation models. This addresses known problems in causality, including the need for domain-specific knowledge when constructing a causal graph and ensuring identifiability of the relevant path-specific effects \citep{Makhlouf2024-causality}. We also account for a known limitation in fair ML literature more broadly, where data are abstracted from the societal context in which they exist \citep{Selbst_fairness_abstraction, Singh2024-sociotechnical}: by integrating domain-specific disparities knowledge, we can ensure that interventions target known problems within the clinical context. 

\textbf{Causal effect estimation in high dimensions requires dimensionality reduction via outcome-driven feature selection.} The underpinning of this fairness pipeline is the ability to robustly estimate path-specific causal effects in high dimensions of $W$, as the dimensionality of the clinical representation ($W$) may vary from hundreds (AMI, SLE, T2DM) to over $1000$ (SCZ). We leverage state-of-the-art doubly robust estimators, which allow for incorrect specification of either the propensity models or the outcome models, and find that for high-dimensional $W$, estimation error is significantly reduced by selecting features based on their predictive power with respect to $Y$, consistent with mediation analyses in high-throughput data \citep{jerolon_group_2024, chen_high-dimensional_2018}. In the case where $Z$ were to be a high-dimensional vector, approaches related to high-dimensional propensity analysis could help ensure robust estimates~\cite{brookhart_variable_2006, schneeweiss_high-dimensional_2009, Zhang2022_lsps}. 

\textbf{Models not explicitly trained for fairness unpredictably introduce causal unfairness.} Across all four cases, we observed a significant increase in the NDE between the dataset and the baseline model (no fairness intervention). The magnitude of the increase varies greatly by task: the NDE increases by factors ranging from 2.4 (T2DM) to 24 (AMI), and the NIE and SE differences between the data and model vary similarly. This result reinforces the importance of context-specific prioritization of causal effects: in addition to identifying the causal pathways through which bias surfaces in the data and the pathways where the causal effects are exacerbated in the baseline model relative to the data, considering the existing healthcare context is essential. This prioritization is also important given the observed tradeoffs between specific causal pathways and model performance. %Beyond examining known sociomedical disparities as previously described, we also recommend examining known disparities in prior healthcare algorithms if possible -- for example, we find that the exacerbation of NIE in the T2DM model is consistent with past (non-ML) clinical risk algorithms. 
Addressing such known limitations of existing clinical risk algorithms may increase the potential for future clinical utility \citep{Chakradeo2025-fairnessconsiderations}. 

\textbf{Consider causal and feature selection-based fairness interventions for reducing path-specific causal effects.} We test interventions from different types of fairness literature, including group fairness (equalized odds \citep{hardt_eqodds}), individual fairness (learning fair representations \citep{pmlr-v28-zemel13}), and causal fairness (path-specific inprocessing \citep{plecko_cfatoolkit_2024}, causal resampling \citep{nabi_learning_2019}). Additionally, we test several naive feature selection approaches not designed to target any specific type of fairness (fairness through unawareness \citep{holtgen_fairnessunawareness}, selection of unbiased features, greedy feature selection). No universally optimal approach exists. Interestingly, we find that causal methods do not always have the best efficacy; relatively naive feature selection approaches, including greedy feature selection and selecting "unbiased" features, can significantly improve causal fairness. We hypothesize that naive feature selection may be a feasible approach in observational health data due to some interdependence between features: even if the "most biased" features are removed from the model, the patterns that underlie a person's health are still learnable from other features in the data. We additionally find that non-causal algorithmic fairness interventions (e.g.. methods that target individual \citep{pmlr-v28-zemel13} and group \citep{hardt_eqodds} fairness objectives) are unable to maintain reasonable model performance when targeting the systemic disparities prevalent in healthcare. These results illustrate the importance of implementing the full pipeline (Figure \ref{fig:workflow}) and trying both causal and non-causal interventions to mitigate path-specific causal effects.

%{\bf{Non-causal algorithmic fairness interventions are ineffective to target systemic disparities prevalent in healthcare.}} The intervention targeting individual fairness, specifically `learning fair representations '~\citep {pmlr-v28-zemel13}, results in near-random performance across all cases (AUROC < 0.55), and the equalized odds~\citep{hardt_eqodds} approach results in near-random performance in all cases except SLE. 

\textbf{Limitations and future work}: We focus on single-attribute fairness tasks. Intersectional fairness approaches may be appropriate for diseases where multiple axes of disparities (e.g., race, gender) are known \citep{Kim2023-ami, Islek2025-amirace}; however, this would require larger datasets with more diverse populations. In the case of T2DM and SCZ, we measure the path-specific effects by comparing White and Black populations; however, both datasets include at least one other race group. While individuals from these groups are included in all models and in the correlation analysis, we do not explicitly model disparities in these other race groups as well. Future work should consider fairness across non-binary categorical variables. Finally, our work explores \textit{tradeoffs} between path-specific causal biases and model performance. While we provide guidance on how to decide which causal pathways to target for bias mitigation, further discussion with clinical experts would be required to select the best model for a given task.
% and our mapping of the SFM does allow for examination of intersectional fairness as long as the sensitive attribute is defined in a binary manner; this could be operationalized by comparing two intersectional subgroups (e.g., Black women and white men) or by comparing one known subgroup to all other individuals (e.g., Asian men vs. non-Asian men and Asian non-men). Such approaches require large amounts of data but constitute important future work.
\section{Conclusion}
In this work, we establish a pipeline for operationalizing path-specific causal fairness in clinical prediction tasks. We define a generalizable mapping of tabular observational health data onto the structural fairness model, and demonstrate how to construct this mapping in the context of known health disparities. We illustrate the importance of defining model fairness through causal pathways that replicate real-world disparity and quantify the extent to which learned models exacerbate these existing biases. By characterizing a more comprehensive fairness-accuracy tradeoff that examines multiple fairness pathways alongside accuracy, we illustrate the importance of context-specific prioritization and enforcement of causal pathways over which a model should be "fair". While our work focuses on the algorithms in the healthcare setting, several components of the pipeline (robust estimation of causal effects, selection of causal pathways to target for bias mitigation, trial of causal and non-causal fairness interventions) are applicable to fair ML outside of healthcare. 

\section{Endmatter}
\subparagraph{\textbf{Ethical Considerations:}} This study was approved by the relevant institutional review board (IRB). Informed consent was waived by the IRB due to low risk to subject welfare and logistical infeasibility of contacting millions of patients from a de-identified database.

\subparagraph{{\textbf{Generative AI usage statement:}}} No AI tools were used to generate text for this manuscript. The authors used generative AI (Gemini 3) to assist with coding (graphing, formatting of tables) and manuscript editing (proofreading, grammar). All generative AI outputs were verified by the authors.

\bibliographystyle{plainnat}
\bibliography{causal_fairness_paper}

\newpage

\section{Path-specific effects for fairness analysis}
\label{appendix:cfa_estimates}
The natural direct effect (NDE) quantifies the impact of changing a person's sensitive attribute while keeping the effects through mediator variables ($W$) at the observed level. Consider the binary sensitive attribute $x \in \{0,1\}$. Then, the direct effect of the sensitive attribute at baseline $x=0$ is given by:

\begin{equation}
\label{eqnde}
NDE_{x_0, x_1}(y) = P(y_{x_1,W_{x_0}} ) - P(y_{x_0})
\end{equation}

The natural indirect effect (NIE) quantifies the change in the outcome distribution mediated through $W$ only. Specifically, the NIE compares the outcome when the mediators' distribution is set to $x=0$ to when the mediators are equal to the values they would be when $x=1$:

\begin{equation}
\label{eqnie}
NIE_{x_1, x_0}(y) = P(y_{x_1,W_{x_0}} ) - P(y_{x_1})
\end{equation}

In the healthcare setting, distinguishing direct and indirect effects is crucial as the NDE  effectively indicates a direct source of discrimination, distinct from the cumulative and broader bias in the healthcare system reflected in the individual's electronic health record (NIE). %The total effect (TE) refers to the combination of the NDE and NIE and is defined as: 

%\begin{equation}
%\label{eqte}
%TE_{x_0, x_1}(y) = P(y_{x_1} ) - P(y_{x_0})
%\end{equation}
Finally, the spurious effect (SE) quantifies the difference between the observational and the interventional outcome distribution. When calculating the NDE and NIE, we move to the interventional setting, where we set the sensitive attribute to $x=0$ or $x=1$ and "cut" the association between $X$ and $Z$. The spurious effect captures the baseline change in the outcome distribution as a result of the two interventions $x=0$ and $x=1$ compared to their observational counterparts. Let, 

\begin{equation}
\label{eqexpsex0}
ExpSE_{x_0}(y) := P(y|x=0) - P(y_{x_0}), \quad   ExpSE_{x_1}(y) := P(y|x=1) - P(y_{x_1})
\end{equation}

The spurious effect is the difference between these two sub-effects, namely: 

\begin{equation}
\label{eqse}
SE = ExpSE_{x_1}(y) - ExpSE_{x_0}(y)
\end{equation}

Thus, a "fair" model with respect to the spurious effect would be one where severing the link between $X$ and $Z$ has the same impact on the outcome $Y$ regardless of the value of $X$.

\section{Robust estimation of path-specific causal effects in synthetic data}
\label{appendix:estimation}

\begin{figure}[htbp]
\centering
\includegraphics[width=1\textwidth]{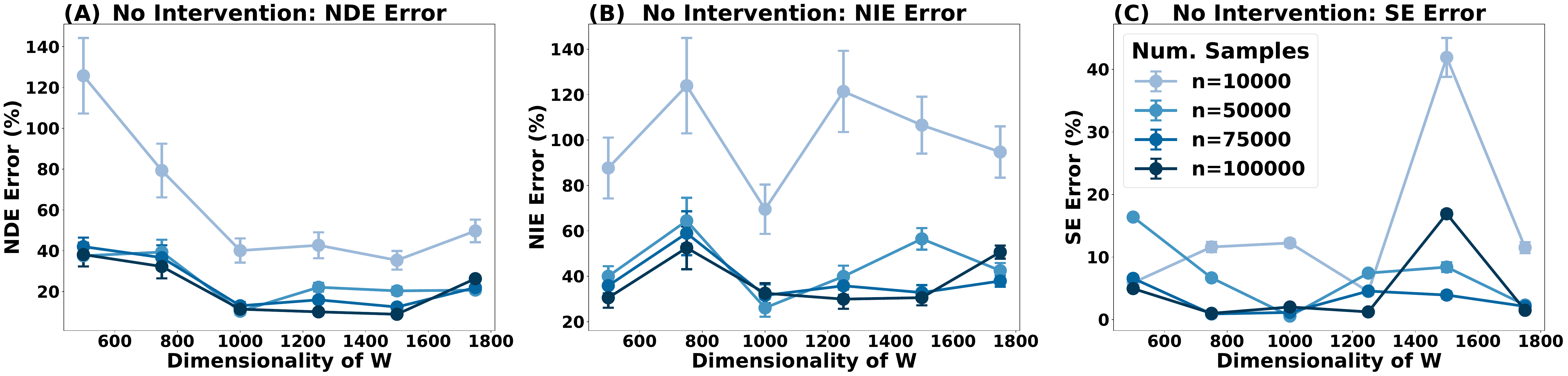}
\caption{Error in the path-specific causal estimates for the (a) natural direct effect (NDE), (b) natural indirect effect (NIE), and (c) spurious effect (SE). Error is highest for a small sample size. The estimates are generated using doubly robust estimation, but still demonstrate large error for NDE and NIE.}
\label{fig:syntheticdata_nointervention}
\Description{Line graph with error bars showing the causal estimation error on the y-axis and the dimensionality of W on the x-axis.}
\end{figure}

We create synthetic data using the causal graph that underlies the SFM (Figure \ref{fig:sfm_dag}a) by separately simulating each edge of the graph as a multilayer perceptron. This allows us to simulate $W$ and $Y$ "naturally" (randomly sampling the protected attribute $X$) and in the counterfactual setting (forcibly setting $X$ to $0$ or $1$ and simulating $W_{X_{0}}$, $W_{X_{1}}$, $Y_{X_{0},W_{X_{0}}}$, $Y_{X_{0},W_{X_{1}}}$, $Y_{X_{0},W_{X_{1}}}$, and $Y_{X_{1},W_{X_{1}}}$). This allows us to calculate the "true" natural direct effect (NDE), natural indirect effect (NIE), and spurious effect (SE) as in \cite{plecko_cfatoolkit_2024}. We select dimensions for simulation of $W$ between $|W| = 750$ and $|W| = 1,750$, as this encapuslates the dimensionality of $W$ in our selected tasks ($|W|=768$ for AMI, SLE, and T2DM; $|W|=1,570$ for SCZ). We select $|Z|=10$ for $|W|=750$ (again, to mimic the dimensionality of the AMI, SLE, and T2DM tasks) and $|Z|=30$ for the other dimensions of $W$ to mimic the SCZ task. The high error in NDE and NIE (Figure \ref{fig:syntheticdata_nointervention}) motivate us to experiment with dimensionality reduction as a method for improving estimate accuracy. 

\begin{table}[ht]
\centering
\resizebox{\textwidth}{!}
{%
\begin{tabular}{cccccccccc}
\toprule
 & & \multicolumn{2}{c}{\textbf{NDE Error (\%)}} & & \multicolumn{2}{c}{\textbf{NIE Error (\%)}} & & \multicolumn{2}{c}{\textbf{SE Error (\%)}} \\
\cmidrule(lr){3-4} \cmidrule(lr){6-7} \cmidrule(lr){9-10}
\textbf{Dim W} & \textbf{N} & \textbf{No Intervention} & \textbf{PFI (20\%)} & & \textbf{No Intervention} & \textbf{PFI (20\%)} & & \textbf{No Intervention} & \textbf{PFI (20\%)} \\
\midrule

\multirow{4}{*}{750} & 10,000 & 79.310 (4.081, 228.536) & 45.571 (2.173, 128.103) & & 123.927 (10.276, 372.327) & 78.161 (3.617, 213.691) & & 11.611 (2.075, 19.487) & 11.611 (2.075, 19.487) \\
 & 50,000 & 39.337 (1.650, 106.857) & 11.398 (0.218, 35.774) & & 64.421 (3.498, 181.584) & 18.071 (0.975, 56.431) & & 6.670 (5.098, 8.307) & 6.670 (5.098, 8.307) \\
 & 75,000 & 36.642 (1.972, 120.840) & 12.232 (0.503, 34.469) & & 58.959 (3.372, 193.431) & 19.884 (0.898, 55.653) & & 0.906 (0.044, 2.328) & 0.906 (0.044, 2.328) \\
 & 100,000 & 32.274 (1.530, 103.406) & 11.406 (0.937, 31.692) & & 52.526 (2.702, 167.299) & 17.664 (1.282, 49.180) & & 1.004 (0.069, 2.220) & 1.004 (0.069, 2.220) \\
\cmidrule{1-10}
\multirow{4}{*}{1,000} & 10,000 & 40.091 (1.598, 111.827) & 21.517 (0.881, 57.915) & & 69.529 (3.591, 193.792) & 40.135 (2.423, 82.119) & & 12.248 (5.856, 17.251) & 12.248 (5.856, 17.251) \\
 & 50,000 & 10.350 (0.508, 28.142) & 13.499 (1.446, 29.392) & & 26.081 (1.064, 69.714) & 29.869 (3.117, 70.365) & & 0.561 (0.035, 1.643) & 0.561 (0.035, 1.643) \\
 & 75,000 & 13.032 (1.150, 38.663) & 4.598 (0.242, 12.138) & & 31.584 (1.855, 96.201) & 15.713 (2.189, 35.134) & & 1.148 (0.257, 2.121) & 1.148 (0.257, 2.121) \\
 & 100,000 & 11.365 (0.251, 32.018) & 4.355 (0.381, 11.531) & & 32.525 (3.196, 86.739) & 18.105 (2.244, 36.530) & & 2.016 (1.047, 2.856) & 2.016 (1.047, 2.856) \\
\cmidrule{1-10}
\multirow{4}{*}{1,250} & 10,000 & 42.681 (2.086, 116.673) & 24.080 (1.929, 53.761) & & 121.390 (6.251, 325.329) & 64.552 (7.563, 141.625) & & 4.476 (0.289, 10.938) & 4.476 (0.289, 10.938) \\
 & 50,000 & 22.106 (2.093, 40.526) & 43.752 (30.457, 54.553) & & 39.915 (3.840, 87.147) & 93.578 (58.198, 125.238) & & 7.433 (4.896, 10.168) & 7.433 (4.896, 10.168) \\
 & 75,000 & 15.936 (2.587, 29.818) & 33.401 (26.190, 40.575) & & 35.799 (1.166, 73.500) & 83.999 (63.064, 105.462) & & 4.560 (2.724, 6.340) & 4.560 (2.724, 6.340) \\
 & 100,000 & 10.053 (0.379, 25.939) & 32.213 (26.405, 37.901) & & 29.930 (1.701, 78.364) & 93.471 (77.614, 111.653) & & 1.233 (0.166, 2.716) & 1.233 (0.166, 2.716) \\
\cmidrule{1-10}
\multirow{4}{*}{1,500} & 10,000 & 35.332 (2.854, 89.697) & 16.419 (1.197, 44.223) & & 106.558 (4.453, 255.723) & 30.348 (1.525, 90.778) & & 41.912 (11.469, 70.110) & 41.912 (11.469, 70.110) \\
 & 50,000 & 20.344 (2.301, 37.949) & 6.178 (0.825, 15.590) & & 56.469 (8.808, 100.389) & 10.657 (0.724, 31.218) & & 8.394 (0.553, 15.522) & 8.394 (0.553, 15.522) \\
 & 75,000 & 12.428 (0.804, 26.085) & 8.365 (2.351, 14.346) & & 32.814 (3.874, 64.966) & 18.465 (4.110, 32.993) & & 3.927 (0.415, 11.430) & 3.927 (0.415, 11.430) \\
 & 100,000 & 8.868 (0.287, 20.068) & 9.830 (3.856, 15.062) & & 30.572 (1.641, 63.498) & 13.857 (1.030, 27.395) & & 16.915 (11.021, 21.612) & 16.915 (11.021, 21.612) \\
\cmidrule{1-10}
\multirow{4}{*}{1,750} & 10,000 & 49.700 (1.787, 107.959) & 20.749 (4.744, 40.926) & & 94.719 (4.689, 203.145) & 35.255 (3.447, 74.458) & & 11.507 (3.299, 19.435) & 11.507 (3.299, 19.435) \\
 & 50,000 & 20.712 (5.537, 35.845) & 4.582 (0.098, 12.260) & & 42.497 (12.590, 72.629) & 9.758 (0.382, 26.426) & & 2.327 (0.242, 4.805) & 2.327 (0.242, 4.805) \\
 & 75,000 & 21.926 (11.099, 34.154) & 6.467 (1.534, 13.016) & & 37.904 (16.668, 62.389) & 7.129 (0.590, 18.568) & & 2.113 (0.541, 3.633) & 2.113 (0.541, 3.633) \\
 & 100,000 & 26.314 (11.594, 38.866) & 7.229 (1.439, 11.650) & & 50.649 (19.963, 77.349) & 12.055 (1.772, 21.804) & & 1.485 (0.105, 2.852) & 1.485 (0.105, 2.852) \\
%\cmidrule{1-10}
\bottomrule
\end{tabular}
}
\caption{Comparison of NDE, NIE, and SE Errors across dimensions and sample sizes. Values shown are Mean percent error (95\% CI). Note that SE is not expected to change, as the representation of $W$ does not impact the spurious effect estimate. PFI ($20\%$) refers to use of permutation feature analysis (PFI) to identify the top $20\%$ of features most important for predicting the outcome $Y$; this dimensionality reduction technique significantly improves the NDE and NIE estimation error in four out of the five experiments shown.}
\label{tab:estimation_error_pfi}
\end{table}

\begin{figure}[htbp]
\centering
\includegraphics[width=1\textwidth]{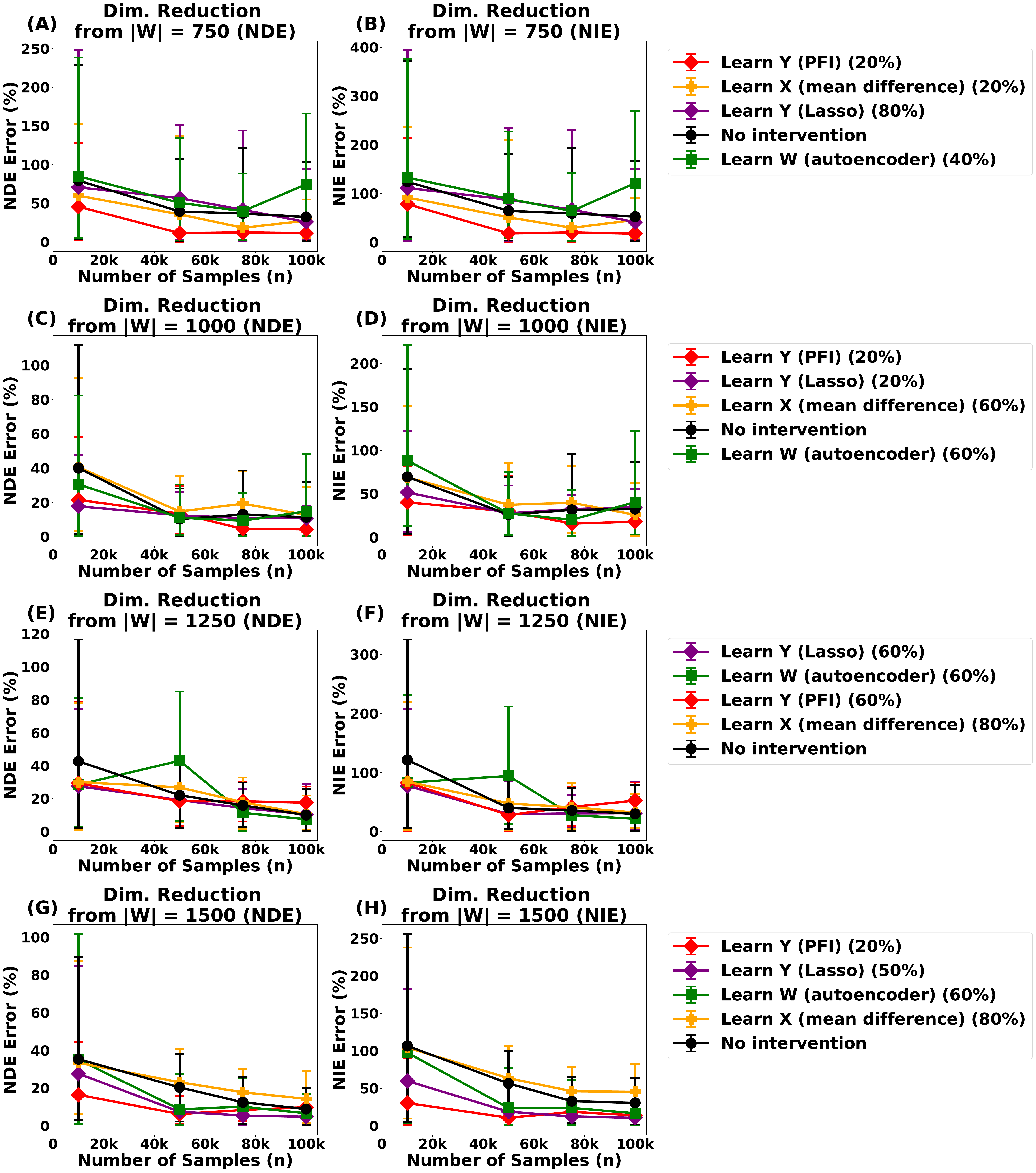}
\caption{Error in the path-specific causal estimates for dimensionality $|W|<1,750$. Each line corresponds to a different dimensionality reduction methods, with the black line corresponding to no dimensionality reduction. The NDE estimation error is reflected in (A), (C), (E), and (G) and the NIE estimation error is reflected in part (B), (D), (F), and (H). We find that interventions that identify features important for predicting the outcome ($Y$) generally outperform other methods.}
\label{fig:syntheticdata_withintervention_750_1500}
\Description{This figure contains a set of 8 subplots -- each row of two subplots visualizes the NDE and NIE error for a given dimension. Each plot is a line plot with colored lines corresponding to the dimensionality reduction method and associated error bars.}
\end{figure}

In addition to testing dimensionality reduction techniques for $|W| = 1,750$ (Figure \ref{fig:syntheticdata_withintervention_1750}), we test dimensionality reduction for $|W|=750$, $|W|=1,000$, $|W|=1,250$, and $|W|=1,500$ (Figure \ref{fig:syntheticdata_withintervention_750_1500}). Across three out of these four synthetic experiments, we find that dimensionality reduction conducted by selecting features important to learning the outcome model ("Learn Y") are most effective lowering estimation error. This finding is consistent with the synthetic experiments for $|W|=1,750$. In Table \ref{tab:estimation_error_pfi}, we present the estimation error under no dimensionality reduction (No Intervention) and the best-performing intervention (using PFI to select the top $20\%$ of important features for predicting $Y$). 

\section{Supplementary Methods}
\label{appendix:methods}
\subsection{Tuning fairness parameters}
In addition to the baseline model, we train models that implement the fairness interventions outlined in Section \ref{sec:fairness_interventions}. In the cases where we train multiple models corresponding to one fairness intervention, we select the model with the best validation-set tradeoff between binary cross entropy loss (performance) and average causal effect (fairness). 

For interventions that impose fairness constraints using a loss regularization term (\textbf{path-specific inprocessing}, \textbf{equalized odds}), the "fairness parameter" ($\lambda$) controls the strength of the loss regularization term relative to the binary cross entropy loss. A larger value for $\lambda$ increases the scale of the loss regularization term; we vary $\lambda$ logarithmically between $0.1$ and $100$. For interventions that use feature selection (\textbf{greedy feature selection}, \textbf{"unbiased" feature selection}), the "fairness parameter" controls the number of features used by the model. For AMI, SLE, and T2DM tasks (where the total dimensionality of $|W|$ is $768$), we test $number\ of\ features=300,\ 400,\ 500,\ 600,\ 700$. For SCZ (total dimensionality of features is $|W| = 1,570$), we test $number\ of\ features=400,\ 500,\ 550,\ 600$. While this corresponds to a lower percentage of features relative to the non-SCZ tasks due to prior experiments that found that limiting the baseline SCZ model to the top $30\%$ of features ($471$) does not diminish model performance. In the case of \textbf{causally fair resampling}, we test the impact of targeting the NDE only or the TE (NDE + NIE). This method does not allow for targeting of the SE. We only report the "better" model due to the similarity of the outputs across all clinical tasks. 

\section{Additional Results: AMI}
\label{appendix:ami_additional_results}

\begin{figure}[htbp]
\centering
\includegraphics[width=1\textwidth]{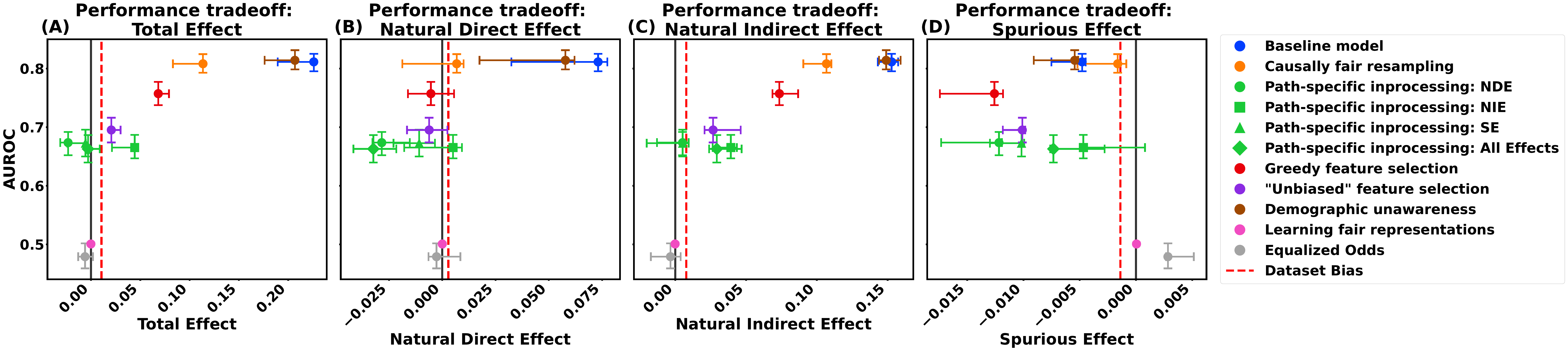}
\caption{AMI: Tradeoff between model performance (AUROC) and path-specific causal effects for each of the tested interventions. Larger AUROC indicates better peformance, and lower magnitude for TE (a), NDE (b), NIE (c), and SE (d) indicate a more "fair" model with respect to that specific causal pathway.}
\label{fig:ami_tradeoffs_plot}
\Description{Set of four subplots with colored points and associated error bars. The x-axis corresponds to the path-specific causal effect and the y-axis corresponds to model performance. The points are colored based on the fairness intervention.}
\end{figure}

The "at-risk" cohort for developing AMI includes individuals with at least two years of observation who have a diagnostic code related to ischemic heart disease. Individuals remain in the cohort if they have had a previous AMI (as this is an acute event), as long as it was over 6 months prior to the prediction visit (to prevent label leakage). We present the tradeoffs between model performance and all path-specific effects in Figure \ref{fig:ami_tradeoffs_plot} and observe the importance of prioritizing causal pathways based on domain knowledge: "unbiased" and greedy feature selection both slightly exacerbate the spurious effect (SE between $-0.015$ and $-0.010$) but demonstrate significant improvement in the NIE ($0.021$ and $0.068$, respectively) relative to causally fair resampling, which eliminates the SE but retains a significantly higher NIE ($0.107$). 

\begin{figure}[htbp]
\centering
\includegraphics[width=0.5\textwidth]{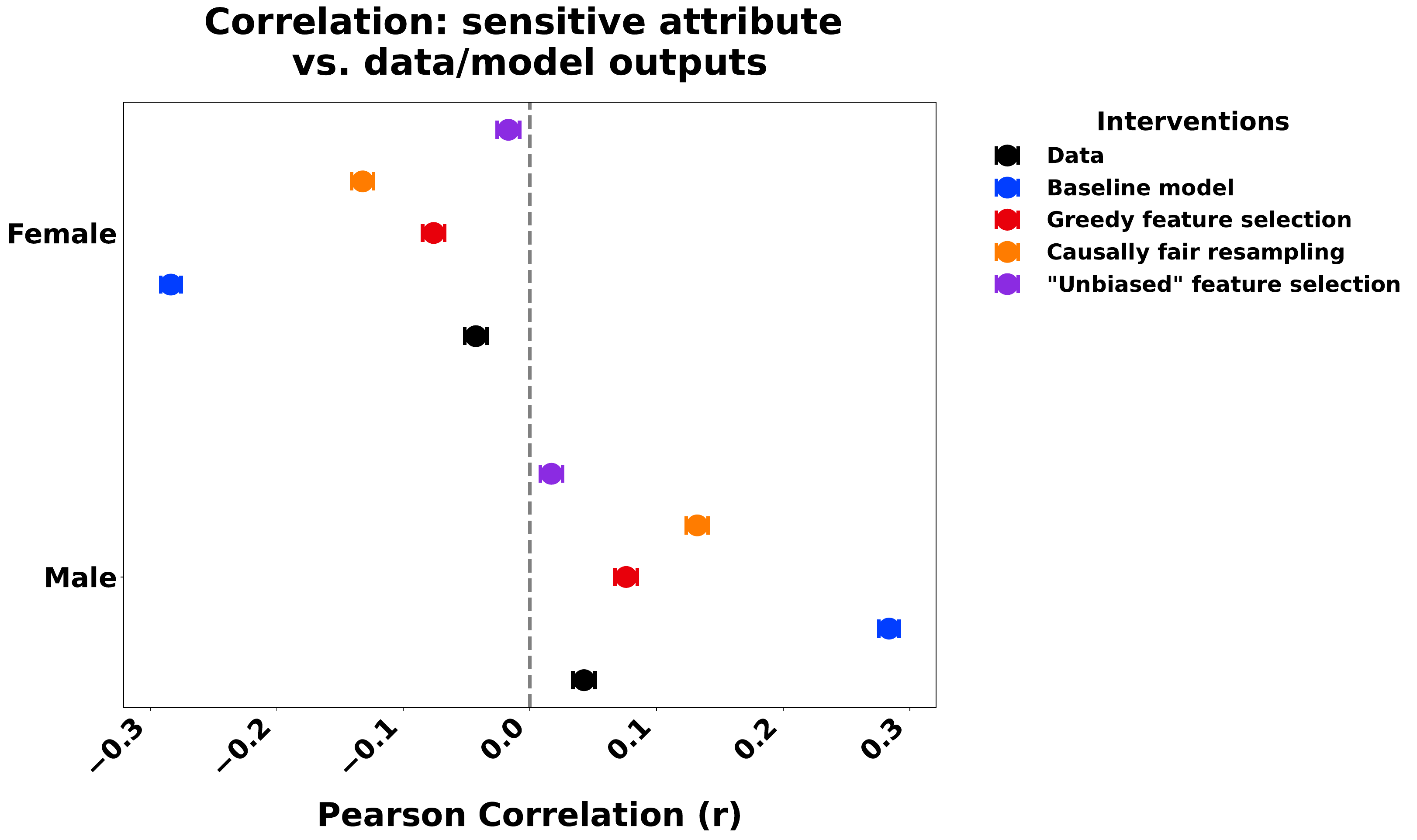}
\caption{Correlation between patient gender and output. We visualize the correlation (Pearson R) between gender and the true output (Data), as well as between gender and the predicted output for the baseline model and best fairness interventions (causally fair resampling, greedy feature selection, "unbiased feature selection).}
\label{fig:ami_correlations}
\Description{Scatter plot with points colored based on the fairness intervention; the x-axis corresponds to Pearson Correlation, and the y-axis corresponds to the sensitive attribute (gender).}
\end{figure}

When we compare the correlations between the sensitive attribute and the (predicted or true) outputs in the models, we find a small positive correlation between male gender and the outcome in the data (Figure \ref{fig:ami_correlations}). This correlation is significantly stronger in the baseline model, and is reduced significantly by each of the three fairness intervention. The ranking of each of gender-prediction correlations over these fairness interventions corresponds to the ranking of NIE, with the highest magnitude NIE and correlations belonging to causally fair resampling, followed by  greedy feature selection and then selection of "unbiased" features. All three interventions, however, lead to significantly lower correlation than exists in the baseline model.

\begin{table}[htbp]
\centering
\label{tab:AMI_results}
\resizebox{\textwidth}{!}{%
\begin{tabular}{lllllll}
\toprule
Experiment & AUROC & Calibration & TE & NDE & NIE & SE \\
\midrule
Baseline model & 0.812 (0.796, 0.825) & 0.387 (0.387, 0.388) & 0.226 (0.189, 0.226) & 0.073 (0.032, 0.077) & 0.153 (0.143, 0.157) & -0.005 (-0.008, -0.004) \\
Data & - & - & 0.011 (0.005, 0.011) & 0.003 (-0.003, 0.004) & 0.008 (0.005, 0.009) & -0.001 (-0.001, -0.000) \\
Causally fair resampling & 0.808 (0.793, 0.825) & 0.011 (0.010, 0.012) & 0.114 (0.083, 0.114) & 0.007 (-0.019, 0.010) & 0.107 (0.090, 0.110) & -0.002 (-0.005, -0.001) \\
Path-specific inprocessing: All Effects & 0.663 (0.640, 0.686) & 0.385 (0.385, 0.386) & -0.003 (-0.003, 0.009) & -0.032 (-0.042, -0.022) & 0.029 (0.024, 0.047) & -0.007 (-0.008, -0.003) \\
Path-specific inprocessing: NDE & 0.674 (0.652, 0.692) & 0.389 (0.389, 0.390) & -0.023 (-0.031, -0.023) & -0.028 (-0.033, -0.015) & 0.005 (-0.013, 0.009) & -0.012 (-0.017, -0.012) \\
Path-specific inprocessing: NIE & 0.665 (0.647, 0.687) & 0.384 (0.384, 0.384) & 0.044 (0.021, 0.045) & 0.005 (-0.018, 0.009) & 0.039 (0.026, 0.044) & -0.005 (-0.005, 0.001) \\
Path-specific inprocessing: SE & 0.673 (0.650, 0.696) & 0.384 (0.384, 0.385) & -0.006 (-0.024, -0.005) & -0.011 (-0.023, -0.003) & 0.005 (-0.020, 0.009) & -0.010 (-0.013, -0.010) \\
Greedy feature selection & 0.757 (0.738, 0.777) & 0.391 (0.391, 0.392) & 0.068 (0.068, 0.079) & -0.005 (-0.016, 0.006) & 0.074 (0.069, 0.087) & -0.013 (-0.017, -0.012) \\
"Unbiased" feature selection & 0.695 (0.674, 0.716) & 0.391 (0.390, 0.391) & 0.021 (0.020, 0.030) & -0.006 (-0.017, 0.002) & 0.027 (0.021, 0.046) & -0.010 (-0.012, -0.010) \\
Demographic unawareness & 0.814 (0.799, 0.832) & 0.389 (0.388, 0.390) & 0.207 (0.176, 0.207) & 0.058 (0.017, 0.062) & 0.149 (0.144, 0.159) & -0.005 (-0.009, -0.005) \\
Learning fair representations & 0.501 (0.500, 0.503) & 0.011 (0.010, 0.012) & -0.000 (-0.000, 0.000) & 0.000 (-0.000, 0.001) & -0.000 (-0.000, 0.000) & 0.000 (-0.000, 0.000) \\
Equalized Odds & 0.479 (0.459, 0.502) & 0.384 (0.384, 0.385) & -0.006 (-0.013, 0.002) & -0.003 (-0.006, 0.009) & -0.003 (-0.017, 0.004) & 0.003 (0.003, 0.005) \\
\bottomrule
\end{tabular}}
\caption{AMI: Impact of fairness interventions on model performance and path-specific causal effects}

\end{table}

\section{Additional Results: SLE}
\label{appendix:sle_additional_results}

\begin{figure}[htbp]
\centering
\includegraphics[width=1\textwidth]{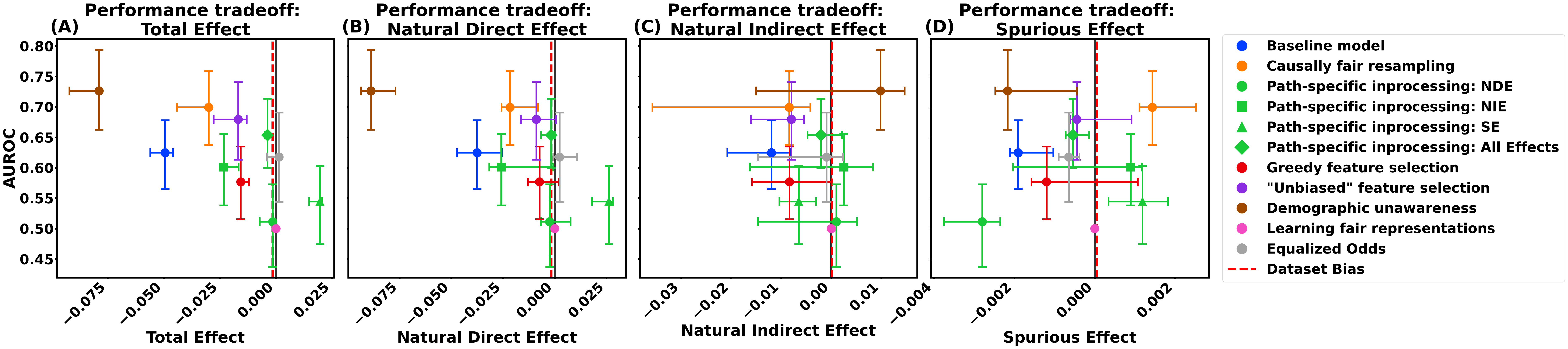}
\caption{SLE: Tradeoff between model performance (AUROC) and path-specific causal effects for each of the tested interventions. Larger AUROC indicates better peformance, and lower magnitude for TE (a), NDE (b), NIE (c), and SE (d) indicate a more "fair" model with respect to that specific causal pathway.}
\label{fig:sle_tradeoffs_plot}
\Description{Set of four subplots with colored points and associated error bars. The x-axis corresponds to the path-specific causal effect and the y-axis corresponds to model performance. The points are colored based on the fairness intervention.}
\end{figure}

\begin{table}[htbp]
\centering
\caption{SLE: Impact of fairness interventions on model performance and path-specific causal effects}
\label{tab:SLE_results}
\resizebox{\textwidth}{!}{%
\begin{tabular}{lllllll}
\toprule
Experiment & AUROC & Calibration & TE & NDE & NIE & SE \\
\midrule
Baseline model & 0.625 (0.565, 0.678) & 0.253 (0.253, 0.253) & -0.050 (-0.056, -0.046) & -0.038 (-0.047, -0.025) & -0.012 (-0.021, -0.008) & -0.002 (-0.002, -0.001) \\
Data & - & - & -0.002 (-0.002, -0.001) & -0.002 (-0.002, -0.001) & 0.000 (-0.000, 0.000) & 0.000 (-0.000, 0.000) \\
Causally fair resampling & 0.699 (0.638, 0.759) & 0.001 (0.001, 0.002) & -0.030 (-0.044, -0.030) & -0.022 (-0.026, -0.008) & -0.008 (-0.036, -0.004) & 0.001 (0.001, 0.003) \\
Path-specific inprocessing: All Effects & 0.654 (0.600, 0.714) & 0.253 (0.253, 0.253) & -0.004 (-0.005, -0.004) & -0.002 (-0.007, 0.000) & -0.002 (-0.005, 0.002) & -0.001 (-0.001, -0.000) \\
Path-specific inprocessing: NDE & 0.511 (0.437, 0.573) & 0.253 (0.253, 0.253) & -0.001 (-0.007, -0.001) & -0.002 (-0.007, 0.008) & 0.001 (-0.015, 0.005) & -0.003 (-0.004, -0.002) \\
Path-specific inprocessing: NIE & 0.601 (0.538, 0.656) & 0.253 (0.253, 0.253) & -0.023 (-0.024, -0.017) & -0.026 (-0.032, -0.001) & 0.002 (-0.016, 0.008) & 0.001 (-0.002, 0.001) \\
Path-specific inprocessing: SE & 0.545 (0.475, 0.603) & 0.253 (0.253, 0.253) & 0.020 (0.015, 0.020) & 0.026 (0.018, 0.028) & -0.007 (-0.010, -0.003) & 0.001 (0.000, 0.002) \\
"Unbiased" feature selection & 0.680 (0.613, 0.741) & 0.253 (0.253, 0.253) & -0.017 (-0.028, -0.013) & -0.009 (-0.016, 0.001) & -0.008 (-0.016, -0.005) & -0.000 (-0.001, 0.001) \\
Greedy feature selection & 0.577 (0.516, 0.635) & 0.253 (0.253, 0.253) & -0.016 (-0.016, -0.012) & -0.007 (-0.013, 0.002) & -0.008 (-0.016, 0.000) & -0.001 (-0.002, 0.001) \\

Demographic unawareness & 0.726 (0.662, 0.794) & 0.253 (0.253, 0.253) & -0.079 (-0.092, -0.079) & -0.089 (-0.094, -0.077) & 0.010 (-0.015, 0.015) & -0.002 (-0.002, -0.000) \\
Learning fair representations & 0.500 (0.500, 0.500) & 0.001 (0.001, 0.002) & -0.000 (-0.000, 0.000) & -0.000 (-0.000, 0.000) & 0.000 (-0.000, 0.000) & 0.000 (-0.000, 0.000) \\
Equalized Odds & 0.618 (0.544, 0.691) & 0.253 (0.253, 0.253) & 0.001 (-0.004, 0.002) & 0.002 (-0.001, 0.011) & -0.001 (-0.015, 0.002) & -0.001 (-0.001, -0.000) \\
\bottomrule
\end{tabular}}
\end{table}

\begin{figure}[htbp]
\centering
\includegraphics[width=0.5\textwidth]{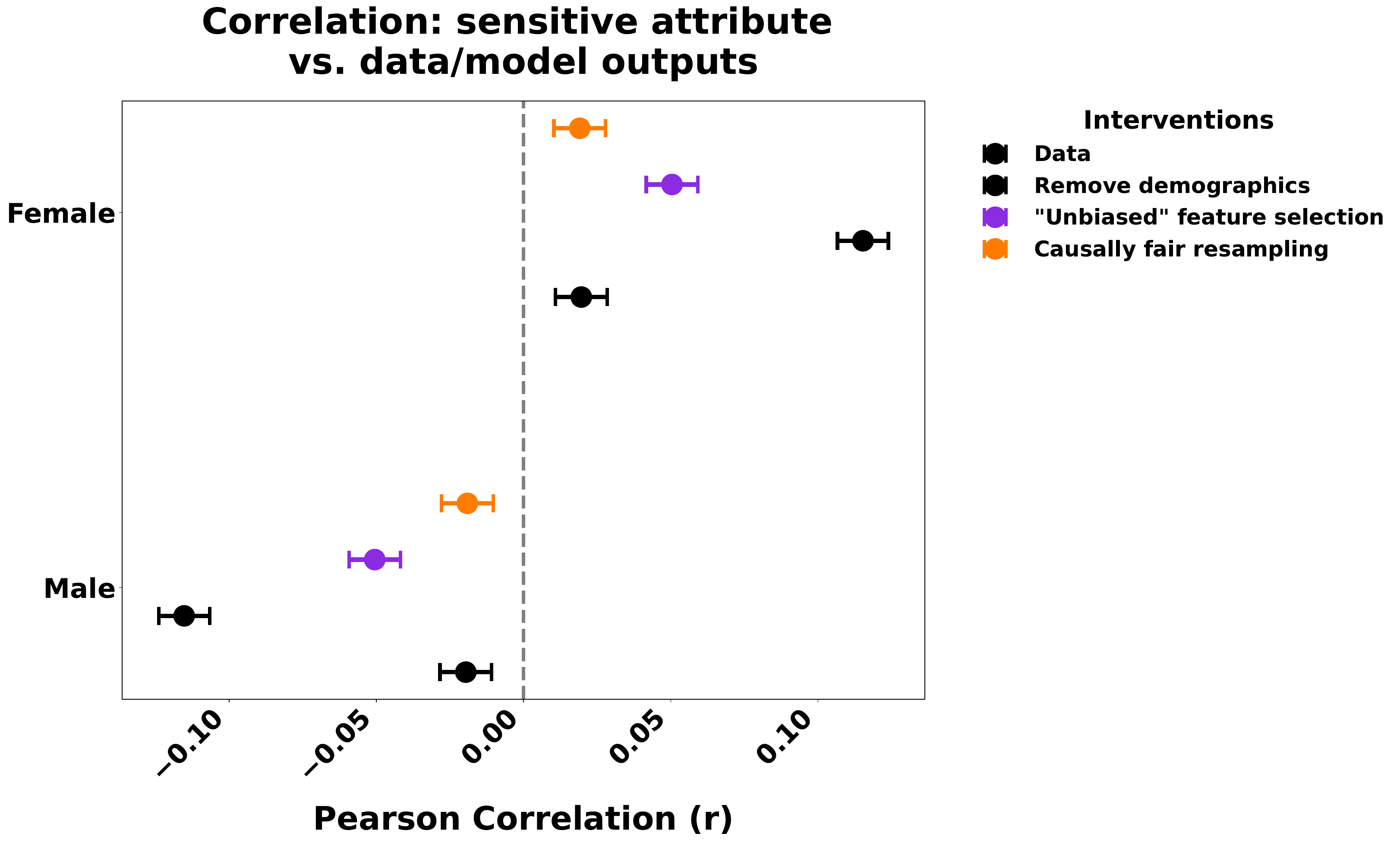}
\caption{Correlation between patient gender and output. We visualize the correlation (Pearson R) between gender and the true output (Data), as well as between gender and the predicted output for the best-performing model (demographic unawareness) and best fairness interventions (causally fair resampling, "unbiased feature selection).}
\label{fig:sle_correlations}
\Description{Scatter plot with points colored based on the fairness intervention; the x-axis corresponds to Pearson Correlation, and the y-axis corresponds to the sensitive attribute (gender).}
\end{figure}

The "at-risk" cohort for SLE comprises individuals with at least one symptom of SLE or one SLE-related prescription. We restrict prediction to the first occurrence of SLE due to its chronic nature. In addition to our findings that causally fair resampling and "unbiased" feature selection achieve the best NDE-performance tradeoff (Figure \ref{fig:sle_tradeoffs_plot}B), we find that path-specific inprocessing methods are able to more effectively control the NIE; however, this comes at a further reduction in model performance. The spurious effect is small ($<0.005$) across all models. 

When we examine correlation between patient gender and (predicted) outcome, we observe the lowest-magnitude correlation between gender and the outcome in the data and the largest correlation in the demographic-unaware model, which exhibited the largest NDE. Causally fair resampling renders the correlation between gender and predicted outcome equal to that in the data, while "unbiased" feature selection demonstrates a slightly larger magnitude of correlation relative to the data (Figure \ref{fig:sle_correlations}).

\section{Additional Results: T2DM}
\label{appendix:t2dm_additional_results}
\begin{figure}[htbp]
\centering
\includegraphics[width=1\textwidth]{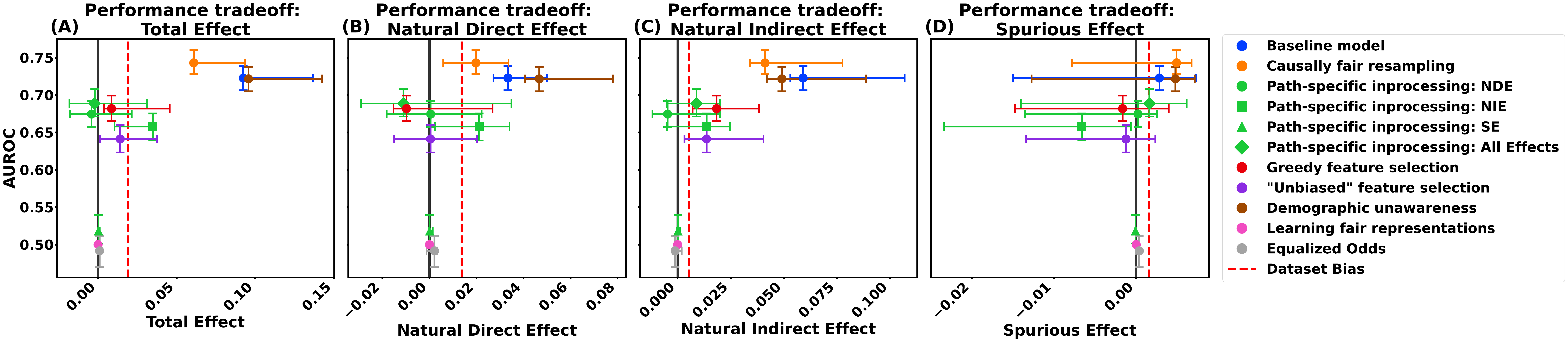}
\caption{T2DM: Tradeoff between model performance (AUROC) and path-specific causal effects for each of the tested interventions. Larger AUROC indicates better peformance, and lower magnitude for TE (a), NDE (b), NIE (c), and SE (d) indicate a more "fair" model with respect to that specific causal pathway.}
\label{fig:t2dm_tradeoffs_plot}
\Description{Set of four subplots with colored points and associated error bars. The x-axis corresponds to the path-specific causal effect and the y-axis corresponds to model performance. The points are colored based on the fairness intervention.}
\end{figure}

\begin{table}[htbp]
\centering
\caption{T2DM: Impact of fairness interventions on model performance and path-specific causal effects}
\label{tab:T2DM_results}
\resizebox{\textwidth}{!}{%
\begin{tabular}{lllllll}
\toprule
Experiment & AUROC & Calibration & TE & NDE & NIE & SE \\
\midrule
Baseline model & 0.723 (0.706, 0.739) & 0.388 (0.387, 0.388) & 0.092 (0.092, 0.137) & 0.033 (0.027, 0.050) & 0.059 (0.053, 0.107) & 0.003 (-0.015, 0.007) \\
Data & - & - & 0.019 (0.003, 0.019) & 0.014 (0.002, 0.016) & 0.005 (-0.004, 0.007) & 0.002 (0.000, 0.002) \\
Causally fair resampling & 0.743 (0.728, 0.760) & 0.017 (0.016, 0.018) & 0.061 (0.060, 0.093) & 0.020 (0.006, 0.034) & 0.041 (0.034, 0.078) & 0.005 (-0.008, 0.007) \\
Path-specific inprocessing: All Effects & 0.689 (0.671, 0.709) & 0.387 (0.387, 0.388) & -0.002 (-0.018, 0.031) & -0.011 (-0.029, 0.035) & 0.009 (-0.005, 0.020) & 0.002 (-0.014, 0.006) \\
Path-specific inprocessing: NDE & 0.675 (0.657, 0.692) & 0.379 (0.378, 0.379) & -0.004 (-0.018, 0.021) & 0.001 (-0.018, 0.022) & -0.005 (-0.012, 0.020) & 0.000 (-0.014, 0.003) \\
Path-specific inprocessing: NIE & 0.658 (0.639, 0.675) & 0.382 (0.382, 0.383) & 0.035 (0.011, 0.035) & 0.021 (0.002, 0.034) & 0.014 (-0.005, 0.025) & -0.007 (-0.023, -0.001) \\
Path-specific inprocessing: SE & 0.518 (0.501, 0.539) & 0.378 (0.378, 0.378) & 0.000 (0.000, 0.001) & 0.000 (-0.000, 0.001) & 0.000 (-0.000, 0.001) & -0.000 (-0.000, -0.000) \\
Greedy feature selection & 0.682 (0.666, 0.699) & 0.379 (0.379, 0.380) & 0.009 (0.004, 0.046) & -0.010 (-0.015, 0.027) & 0.018 (0.007, 0.038) & -0.002 (-0.015, 0.004) \\
"Unbiased" feature selection & 0.641 (0.623, 0.660) & 0.383 (0.383, 0.384) & 0.014 (0.001, 0.037) & 0.000 (-0.015, 0.020) & 0.014 (0.003, 0.040) & -0.001 (-0.013, 0.002) \\
Demographic unawareness & 0.722 (0.705, 0.737) & 0.387 (0.386, 0.388) & 0.096 (0.095, 0.142) & 0.047 (0.041, 0.078) & 0.049 (0.042, 0.089) & 0.005 (-0.013, 0.007) \\
Learning fair representations & 0.500 (0.500, 0.500) & 0.017 (0.016, 0.018) & -0.000 (-0.000, 0.000) & -0.000 (-0.000, 0.000) & 0.000 (-0.000, 0.000) & 0.000 (-0.000, 0.000) \\
Equalized Odds & 0.492 (0.470, 0.511) & 0.382 (0.381, 0.382) & 0.001 (0.001, 0.002) & 0.002 (-0.001, 0.003) & -0.001 (-0.002, 0.002) & 0.000 (-0.000, 0.000) \\
\bottomrule
\end{tabular}}
\end{table}

\begin{figure}[htbp]
\centering
\includegraphics[width=0.5\textwidth]{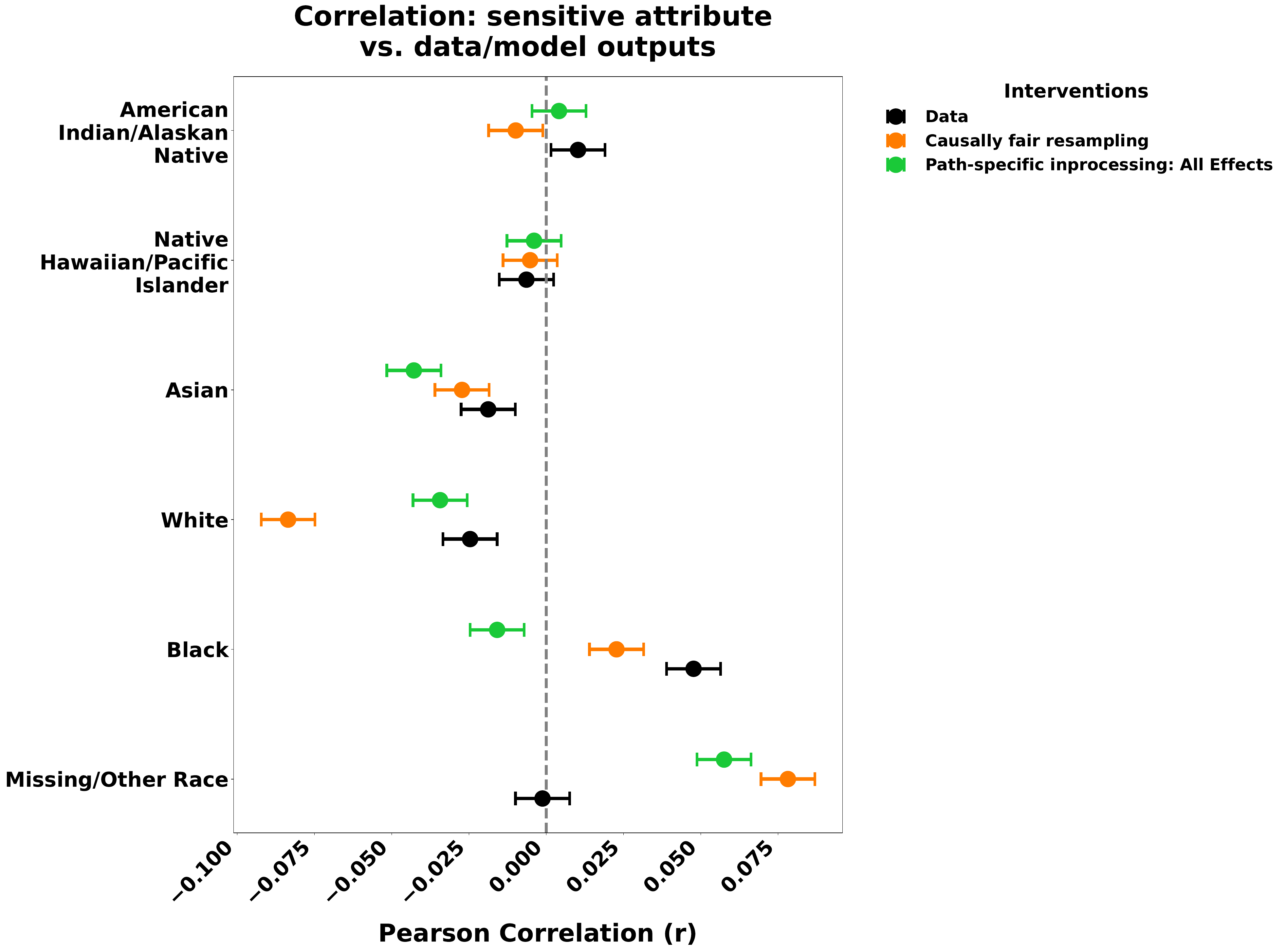}
\caption{Correlation between patient race and output. We visualize the correlation (Pearson R) between race and the true output (Data), as well as between race and the predicted output for the best-performing model (causally fair resampling) and best fairness interventions (path-specific inprocessing: all effects).}
\label{fig:t2dm_correlations}
\Description{Scatter plot with points colored based on the fairness intervention; the x-axis corresponds to Pearson Correlation, and the y-axis corresponds to the sensitive attribute (race).}
\end{figure}

We use a validated diabetes phenotype definition \citep{Suchard2021LEGENDT2DM} to identify individuals from our "at risk" cohort who go on to develop T2DM. Prediction is restricted to the first diagnosis of T2DM due to its chronic nature. Beyond the NDE-NIE-performance tradeoff discussed in Section \ref{sec:fairness_intervention_results}, we find that the spurious effect is insignificant in the baseline model and across all fairness interventions (Figure \ref{fig:t2dm_tradeoffs_plot}D). We now examine the correlation between the sensitive attribute (race) and true/predict outcome across the data and models (Figure \ref{fig:t2dm_correlations}). For White individuals, the causally fair resampling intervention (highest performance, largest NDE and NIE) demonstrates a strong negative correlation between race and predicted outcome. The data and the path-specific in-processing intervention (lower performance, insignificant NDE and NIE) demonstrate lower-magnitude negative correlations. For Black individuals, both the causally fair resampling model and the path-specific in-processing model are associated with lower-magnitude correlations between race and predicted output compared to the raw data; however, this correlation is positive for causally fair resampling ($0.023,\ [0.014,\ 0.031]$) and negative for path-specific in-processing ($-0.016,\ [-0.025,\ -0.007]$). For American Indian/Alaskan Native and Native Hawaiian/Pacific Islander individuals, the correlations are similarly small in the data and both models. There is a negative correlation between race and T2DM for Asian individuals, but this correlation is again similar across the data and both models. For individuals with race recorded as Missing/Other, we find no correlation between race and T2DM in the data, but a significant positive correlation in both models; this correlation is stronger under the causally fair resampling intervention.

\section{Additional Results: SCZ}
\label{appendix:scz_additional_results}
\begin{figure}[htbp]
\centering
\includegraphics[width=1\textwidth]{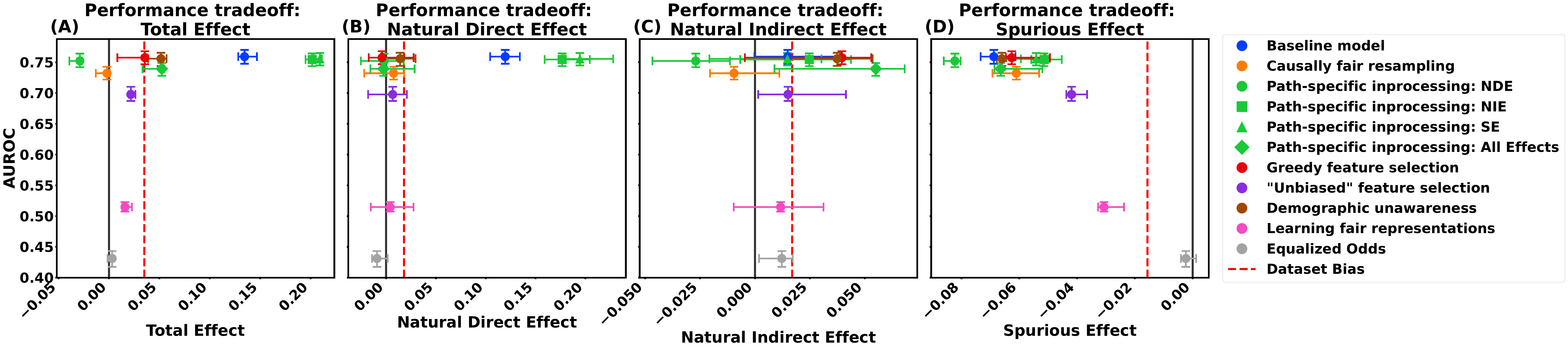}
\caption{SCZ: Tradeoff between model performance (AUROC) and path-specific causal effects for each of the tested interventions. Larger AUROC indicates better peformance, and lower magnitude for TE (a), NDE (b), NIE (c), and SE (d) indicate a more "fair" model with respect to that specific causal pathway.}
\label{fig:scz_tradeoffs_plot}
\Description{Set of four subplots with colored points and associated error bars. The x-axis corresponds to the path-specific causal effect and the y-axis corresponds to model performance. The points are colored based on the fairness intervention.}
\end{figure}

\begin{table}[htbp]
\centering
\caption{SCZ: Impact of fairness interventions on model performance and path-specific causal effects}
\label{tab:SCZ_results}
\resizebox{\textwidth}{!}{%
\begin{tabular}{lllllll}
\toprule
Experiment & AUROC & Calibration & TE & NDE & NIE & SE \\
\midrule
Baseline model & 0.759 (0.748, 0.770) & 0.065 (0.063, 0.068) & 0.134 (0.128, 0.147) & 0.120 (0.104, 0.134) & 0.015 (-0.000, 0.039) & -0.069 (-0.073, -0.061) \\
Data & - & - & 0.035 (0.029, 0.036) & 0.018 (0.005, 0.031) & 0.017 (0.003, 0.027) & -0.016 (-0.017, -0.012) \\
Causally fair resampling & 0.732 (0.722, 0.743) & 0.066 (0.064, 0.069) & -0.002 (-0.013, 0.001) & 0.008 (-0.022, 0.018) & -0.010 (-0.020, 0.011) & -0.061 (-0.069, -0.053) \\
Path-specific inprocessing: All Effects & 0.739 (0.728, 0.749) & 0.066 (0.064, 0.069) & 0.052 (0.033, 0.055) & -0.003 (-0.016, 0.029) & 0.055 (0.009, 0.068) & -0.066 (-0.069, -0.052) \\
Path-specific inprocessing: NDE & 0.752 (0.742, 0.764) & 0.067 (0.064, 0.069) & -0.029 (-0.039, -0.027) & -0.002 (-0.025, 0.018) & -0.027 (-0.047, -0.011) & -0.082 (-0.086, -0.080) \\
Path-specific inprocessing: NIE & 0.755 (0.744, 0.764) & 0.066 (0.063, 0.068) & 0.202 (0.195, 0.204) & 0.177 (0.159, 0.204) & 0.025 (-0.007, 0.044) & -0.051 (-0.056, -0.045) \\
Path-specific inprocessing: SE & 0.755 (0.745, 0.765) & 0.066 (0.063, 0.068) & 0.209 (0.203, 0.211) & 0.194 (0.179, 0.228) & 0.015 (-0.021, 0.030) & -0.054 (-0.059, -0.050) \\
Greedy feature selection & 0.758 (0.747, 0.768) & 0.065 (0.063, 0.068) & 0.035 (0.008, 0.037) & -0.004 (-0.017, 0.028) & 0.040 (-0.005, 0.053) & -0.063 (-0.065, -0.049) \\
"Unbiased" feature selection & 0.698 (0.687, 0.710) & 0.068 (0.065, 0.070) & 0.022 (0.020, 0.026) & 0.007 (-0.018, 0.021) & 0.015 (0.001, 0.041) & -0.042 (-0.044, -0.037) \\
Demographic unawareness & 0.756 (0.744, 0.765) & 0.065 (0.063, 0.068) & 0.052 (0.039, 0.057) & 0.014 (-0.004, 0.030) & 0.037 (0.016, 0.054) & -0.066 (-0.068, -0.051) \\
Learning fair representations & 0.515 (0.507, 0.523) & 0.071 (0.069, 0.074) & 0.016 (0.014, 0.023) & 0.004 (-0.015, 0.028) & 0.012 (-0.010, 0.031) & -0.031 (-0.033, -0.024) \\
Equalized Odds & 0.431 (0.417, 0.443) & 0.071 (0.069, 0.074) & 0.003 (-0.002, 0.004) & -0.009 (-0.014, 0.002) & 0.012 (0.002, 0.017) & -0.002 (-0.004, 0.001) \\
\bottomrule
\end{tabular}}
\end{table}

\begin{figure}[htbp]
\centering
\includegraphics[width=0.5\textwidth]{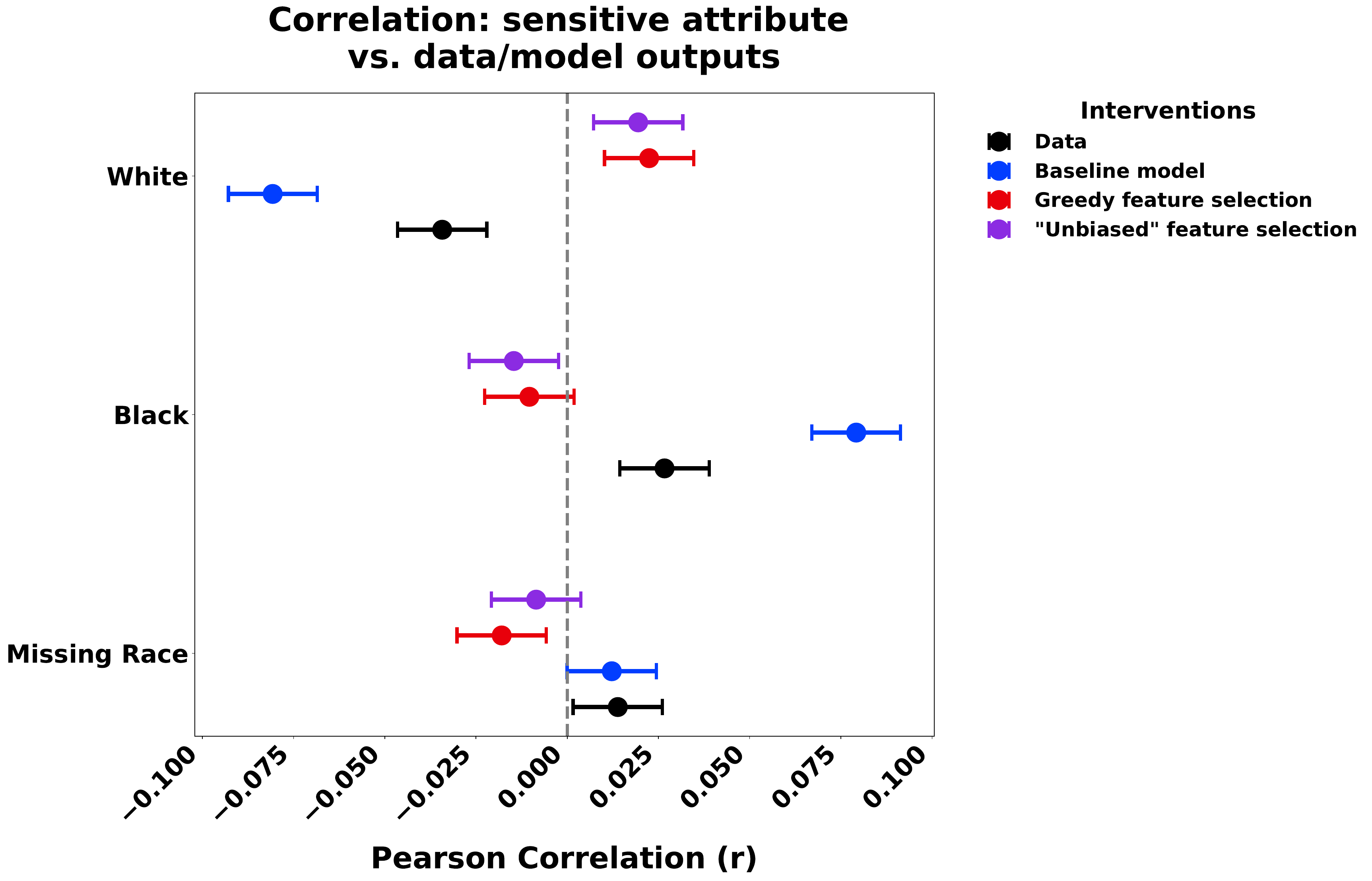}
\caption{Correlation between patient race and output. We visualize the correlation (Pearson R) between race and the true output (Data), as well as between race and the predicted output for the baseline model and best fairness interventions (greedy feature selection, "unbiased" feature selection).}
\label{fig:scz_correlations}
\Description{Scatter plot with points colored based on the fairness intervention; the x-axis corresponds to Pearson Correlation, and the y-axis corresponds to the sensitive attribute (race).}
\end{figure}

 We define our schizophrenia cohort as individuals with at least two SCZ diagnoses and three years of observation prior to the first diagnosis, as validated in \citet{finnerty_prevalence_2024}. We made our prediction (any future diagnosis of SCZ) one year after an individual's first psychosis diagnosis and removed individuals who were diagnosed with SCZ within one year of their initial psychosis diagnosis. We focus on SCZ prediction in the MDCD dataset because there is not sufficient power for these analyses in the CUMC-EHR dataset. When we perform fairness interventions on the schizophrenia model, we observe clear tradeoffs between fairness with respect to the NDE and the NIE: demographic awareness and greedy feature selection both significantly reduce the NDE, but they increase the NIE (Figure \ref{fig:scz_tradeoffs_plot}). When we examine the correlation between race and true/predicted output (Figure \ref{fig:scz_correlations}), the baseline model exacerbates this correlation relative to the data for Black and White patients. Both the greedy feature selection and "unbiased" feature selection result in similar correlations, which lower the magnitude of the correlation to the level found in the data. For patients with missing race, the correlation between race and output is similar in the data and baseline model and is reversed, although similar in magnitude in both the "fairer" models. All correlations between "Missing" race and SCZ output are small ($<0.015$). 

\end{document}